\documentclass{article}

\pdfoutput=1

\usepackage{url}
\usepackage[nonatbib,preprint]{SLI_2023}

\usepackage{graphicx}
\usepackage{booktabs}
\usepackage{amsfonts}       
\usepackage{nicefrac}
\usepackage{microtype}      
\usepackage{xcolor}

\usepackage{pifont}
\newcommand{\cmark}{\ding{51}}%
\newcommand{\xmark}{\ding{55}}%
\usepackage{mathabx}
\usepackage{hyperref}
\usepackage{multirow}
\usepackage{enumitem}
\usepackage{tipa} 
\usepackage{makecell}
\usepackage{tabularx}
\usepackage{siunitx,booktabs}
\usepackage[T1]{fontenc}

\usepackage{soul}
\usepackage{tabularx,colortbl}

\newcommand{\hlc}[2][yellow]{{%
    \colorlet{foo}{#1}%
    \sethlcolor{foo}\hl{#2}}%
}

\newcolumntype{L}{>{\ttfamily}l}

\definecolor{COLORA}{rgb}{0.33999999999999997, 0.86, 0.3712}
\definecolor{COLORB}{rgb}{0.33999999999999997, 0.8287999999999999, 0.86}
\definecolor{COLORC}{rgb}{0.8287999999999999, 0.86, 0.33999999999999997}
\definecolor{COLORD}{rgb}{0.86, 0.33999999999999997, 0.8287999999999999}
\definecolor{COLORE}{rgb}{0.3712, 0.33999999999999997, 0.86}
\definecolor{COLORF}{rgb}{0.86, 0.3712, 0.33999999999999997}
\definecolor{COLORG}{rgb}{0.5, 0.5, 0.5}
\definecolor{color_pron}{HTML}{1D5D9B}
\definecolor{color_conc}{HTML}{7A316F}

\definecolor{niceblueshade}{HTML}{0C6DC7}   
\definecolor{colorknowledge}{HTML}{1F5B93}  
\definecolor{niceblue}{HTML}{1F5B93}   
\definecolor{nicered}{HTML}{BE533B}  
\definecolor{nicegreen}{HTML}{54AD72}  
\definecolor{nicegray}{rgb}{0.3, 0.3, 0.3}

\definecolor{DD}{HTML}{ff165d}   
\definecolor{CC}{HTML}{ff9a00}  
\definecolor{AA}{HTML}{15b7b9}  
\definecolor{BB}{HTML}{6a2c70} 
\newcommand{\PatternA}{Correct reasoning step}
\newcommand{\PatternB}{Correct recall of knowledge}
\newcommand{\PatternC}{Correct reading comprehension}
\newcommand{\PatternD}{Incorrect reasoning step}
\newcommand{\PatternE}{Incorrect or insufficient knowledge}
\newcommand{\PatternF}{Incorrect reading comprehension}

\newcommand\pmx[1]{\raisebox{0.1em}{\color{nicegray}\scriptsize{$\pm$#1}}}

\title{Spoken Language Intelligence of Large Language Models for Language Learning}

\author{%
  Linkai Peng \\
  NetEase Youdao\\
  Beijing 100193, China \\
  \texttt{penglinkai96@gmail.com} \\
  \And
  Baorian Nuchged \\
  The University of Texas at Austin \\
  Austin, TX 78712 \\
  \texttt{baorian@utexas.edu} \\
  \And
  Yingming Gao \\
  Beijing University of Posts and Telecommunications \\
  Beijing 100876, China \\
  \texttt{yingming.gao@bupt.edu.cn} \\
}

\begin{document}

\maketitle

\begin{abstract}

People have long hoped for a conversational system that can assist in real-life situations, and recent progress on large language models (LLMs) is bringing this idea closer to reality. While LLMs are often impressive in performance, their efficacy in real-world scenarios that demand expert knowledge remains unclear. LLMs are believed to hold the most potential and value in education, especially in the development of Artificial intelligence (AI) based virtual teachers capable of facilitating language learning. Our focus is centered on evaluating the efficacy of LLMs in the realm of education, specifically in the areas of spoken language learning which encompass phonetics, phonology, and second language acquisition. 
We introduce a new multiple-choice question dataset to evaluate the effectiveness of LLMs in the aforementioned scenarios, including understanding and application of spoken language knowledge. In addition, we investigate the influence of various prompting techniques such as zero- and few-shot method (prepending the question with question-answer exemplars), chain-of-thought (CoT, think step-by-step), in-domain exampler and external tools (Google, Wikipedia). We conducted large-scale evaluation on popular LLMs (20 distinct models) using these methods. We achieved significant performance improvements compared to the zero-shot baseline in the practical questions reasoning (GPT-3.5, 49.1\% -> 63.1\%; LLaMA2-70B-Chat, 42.2\% -> 48.6\%). We found that models of different sizes have good understanding of concepts in phonetics, phonology, and second language acquisition, but show limitations in reasoning for real-world problems. Additionally, we also explore preliminary findings on conversational communication.
\end{abstract}

\section{Introducation}

The highly parallelizable Transformer architecture \cite{vaswani2017attention} with massively parallel computation hardware and the self-supervised learning technique promise to leverage the vast quantity of raw data (e.g., text, images, audio) to learn general-purpose deep contextualized representations \cite{vaswani2017attention,doersch2015unsupervised,baevski2020wav2vec}. These pre-trained context-aware representations are now ubiquitous in natural language processing and very effective as general-purpose semantic features, which have largely raised the performance of natural language processing (NLP) tasks.

\paragraph{Large Language Models (LLMs)} Language models have revolutionized NLP in recent years. Researchers find that enlarging Pre-trained Languge Model (PLM) (e.g., model parameters) often leads to better performance \cite{hoffmann2022training,kaplan2020scaling}. A number of studies have explored to push the limit of performance by training an ever larger PLM (e.g., the 175B-parameter GPT-3 \cite{brown2020language} and the 540B parameter PaLM \cite{chowdhery2022palm}). A remarkable success of LLMs is ChatGPT\footnote{https://openai.com/blog/chatgpt/}, developed by OpenAI, that adapts the LLMs from the GPT series for dialogue, which presents an amazing conversation ability with humans, which triggered a community-wide enthusiasm to build various imaginative and fantastic applications. Microsoft AI scientists have recently explored the capabilities of OpenAI's GPT-4 \cite{openai2023gpt4}, a more powerful large language model and claimed that GPT-4 demonstrates “sparks” of human-level intelligence, or artificial general intelligence (AGI) \cite{bubeck2023sparks}. 
There have been a number of research efforts aiming at evaluating ChatGPT and other LLMs from different aspects, encompassing a range of factors such as natural language tasks, reasoning, robustness, trustworthiness, ethical considerations and some sepecific applications (e.g. natural science, social science, engineering, medical application, education, etc.) \cite{bang2023multitask,hendrycks2020measuring,castro2023large,frank2023baby,valmeekam2022large,lievin2022can,dai2023can}. The broad applicability of LLMs underscores the evaluation of emergent intelligence in expert domain.

\paragraph{Spoken Language Intelligence (SLI)}\label{para:SLI} Our world is inherently multimodal, and we engage with our environment through a variety of mediums, including text, images, sounds, and sensory experiences. There exist numerous multimodal methods that exhibit exceptional problem-solving capabilities across various scenarios, with text-based semantics frequently serving as the key agent, either explicitly or implicitly \cite{baltruvsaitis2018multimodal,xu2023multimodal}. For example, we aspire to empower machines to comprehend an image or video and articulate its meaning in words \cite{hossain2019comprehensive}. Conversely, we create specialized text prompts to generate images with the creativity of professional painters \cite{rombach2022high}. In the realm of speech, Automatic Speech recognition (ASR) technology is capable of extracting corresponding text from speech signals even in complex speaking scenarios. Conversely, Text-to-Speech (TTS) systems can realistically produce speech sounds that correspond to a given piece of text, mimicking the nuances of human speech. The powerful capabilities of LLM make it easy to integrate these systems together, for example, to build a voice assistants. Owing to the unified Transformer structure, in addition to simple cascading, they can also be completely integrated in an end-to-end manner \cite{ling2023adapting}. LLMs have the ability to replace the LM module in ASR, decode discretized representations directly, or even replace the intricate TTS text frontend, resulting in more expressive speech generation \cite{sigurgeirsson2023using}. Recently, AudioPaLM \cite{rubenstein2023audiopalm} fuses text-based and speech-based language models to inherit the capability to preserve paralinguistic information and intonation from AudioLM \cite{borsos2023audiolm} and the linguistic knowledge present in text LLMs of PaLM-2 \cite{anil2023palm}. It outperforms existing systems for speech translation tasks and has the ability to perform \texttt{zero-shot} speech-to-text translation for many languages. We acknowledged that LLMs have the potential to exhibit SLI; however, their capabilities are still in question whether it can match the expertise of phonetics specialists or serve as effective language learning assessors. To begin exploring this question, we propose a challenging sub-question: Do LLMs possess adequate Spoken Language Intelligence to perform reasoning questions that require the expertise of human phonetic professionals?

\paragraph{Spoken Language Learning} Spoken language learning is the process of acquiring the ability to communicate verbally in a new language. It involves developing skills such as listening, speaking, pronunciation, vocabulary, grammar, and discourse \cite{eskenazi2009overview}. Spoken language learning can be accomplished through diverse methods, including formal instruction, self-study, immersion, interaction, and technology. One of the approaches to integrating technologies into spoken language learning is computer-assisted pronunciation training is computer-assisted pronunciation training (CAPT), a subfield of computer-assisted language training (CALL) \cite{rogerson2021computer,kang2017assessment}. It mainly involves assessment of pronunciation errors, detection and correction of prosody errors. Most current systems focus on the pronunciation of phonemes. Evaluating prosody has always been a challenging task since precise prosody can only be comprehended by grasping the context, which was beyond the capability of previous models \cite{kang2010suprasegmental}. To help language learners develop their spoken language skills, an open and unconstrained\footnote{The content for pronunciation training does not need to be pre-set. Users are free to choose and switch between topics at their leisure. What we are discussing here is not the ethical problem and bias of LLMs.} language learning system is necessary, allowing them to freely express themselves. LLMs are poised to make this dream a reality with the ability to evaluate the emotions, prosody, and even rhythm of a sentence in any given context.

\paragraph{Prompting Engineering} Prompting provides a natural and intuitive interface for humans to interact with LLMs, allowing users to design and supply tailored prompts to guide LLMs in generating desired responses or completing specific tasks. A typical prompting method is \texttt{In-Context Learning} (ICL)  \cite{brown2020language}, which utilizes natural language text to formulates the task description and/or demonstrations, enabling LLMs to recognize and perform a new task by learning from a few given examples. Furthermore, to further improve ICL, \texttt{chain-of-thought} (CoT) prompting  \cite{wei2022chain} incorporate a sequence of intermediate reasoning steps into prompts. Instead of simply constructing the prompts with input-output pairs as in ICL, CoT incorporates intermediate reasoning steps that guide the LLM's reasoning process from the input to the final output within the prompts. Once appropriately designed, CoT prompts can effectively stimulate the reasoning skills of LLMs. In fact, using diverse CoTs (i.e., sample multiple reasoning paths for one problem) has been shown to be a direct and effective approach to enhancing their performance \cite{wang2022self}. Moreover, LLMs are not limited to their internal knowledge and can use external tools when needed. Previous research has demonstrated the use of API calls to integrate various tools, such as search engines, calculators, and compilers, improves the performance of LLMs on specific tasks \cite{nakano2021webgpt,schick2023toolformer,gao2023pal}.

\paragraph{Deploying LLMs} To apply LLMs to real-life scenarios, additional safeguards should be implemented. Language models may amplify social biases present in their training data, may generate incorrect information \cite{Bender2021-fv}. In reality, similar issues have been observed in various language evaluation scenarios, such as native speaker judgments of non-native speakers' speech, which are notoriously biased. This phenomenon is known as \texttt{reverse linguistic stereotyping}, whereby a person's speaking performance is evaluated based on the stereotypes associated with their social identity \cite{kang2009reverse}. Therefore, deploying LLMs in sensitive areas such as education must be approached with great care \cite{KASNECI2023102274,0000198}. While LLMs have been tested on large benchmarks, such as MMLU~ \cite{hendrycks2020measuring} and BIG-bench~ \cite{Srivastava2022-jl}, further studies are necessary to apply them to the domain of spoken language learning.

\paragraph{Contributions}
This paper investigates the performances, interpretability, and limitations of prompting methods in spoken language question-answering. We have collected a composite dataset that includes a set of concepts mainly designed to test the large models' knowledge of spoken language and application questions toward industrial production. We utilized the GPT \cite{openai2023gpt4}, LLaMA \cite{touvron2023llama,touvron2023llama2}, FLAN-T5 \cite{chung2022scaling}, UL2 \cite{tay2022unifying} and Pythia \cite{biderman2023pythia} series for our research, which is conducted in two rounds. We comprehensively review their performance on a large scale and consider two prompting strategies: direct and CoT prompting, under both zero-shot and few-shot learning paradigms. In the second round, we meticulously analyze representative models with advanced optimized methods. The main contributions of this paper are:

\begin{itemize}
    \item We introduced a dataset about spoken language intelligence that serves as a substantial benchmark for speech and language learning scenarios, providing valuable supplementation to existing benchmarks.
    \item We conducted a study on various prompt variations (such as zero-shot, few-shot, direct/CoT, domain-specific sampler, and external tools) and analyze their performance in multiple-choice question form.
    \item We demonstrated that in-domain example sampling techniques can consistently improve performance on domain-specific data.
    \item We conducted an expert evaluation of a small set of multi-turn conversation generated by GPT-3.5\footnote{Unless otherwise specified, the results mentioned in this paper regarding GPT-3.5 are based on the \texttt{GPT-3.5-turbo-0613} model.}. Error analysis indicates that GPT-3.5 has potential to enhance conversational spoken language learning.
\end{itemize}

\section{Dataset}

\begin{figure}[h]
  \centering
  \includegraphics[width=1\linewidth]{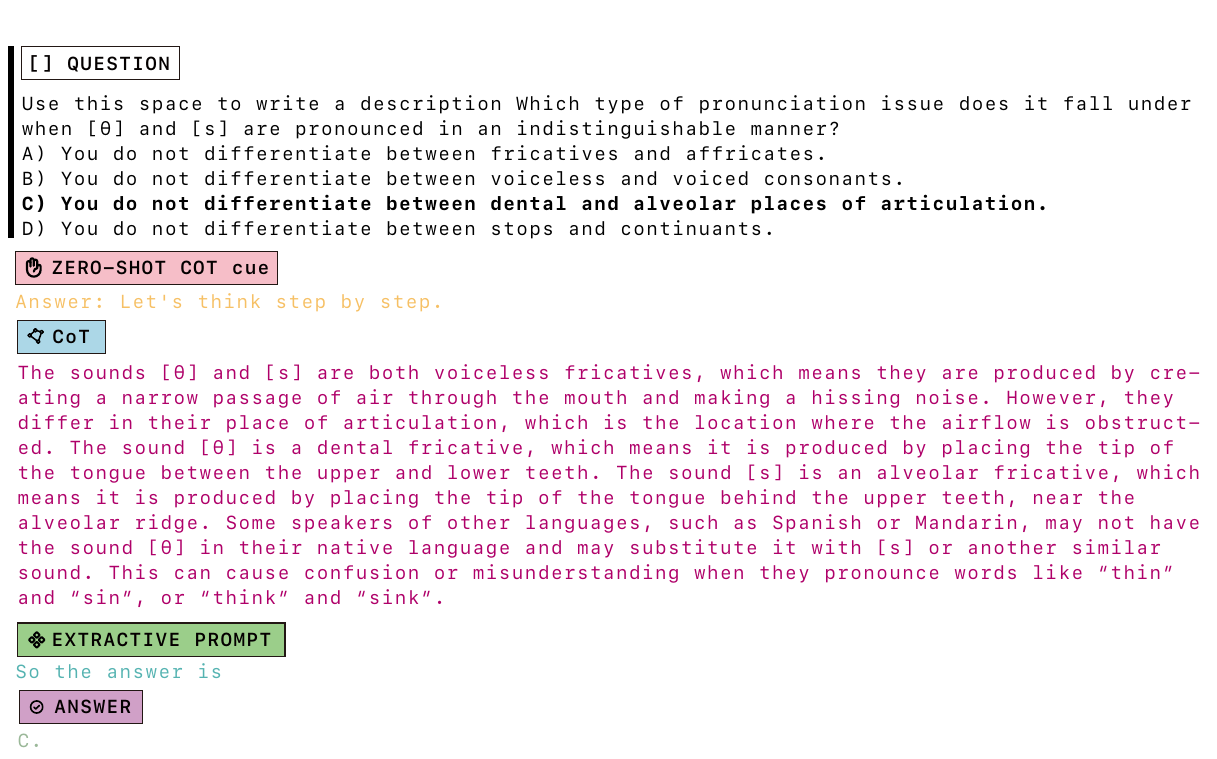}
  \caption{An example of answering a SLIQ-LL Application Question using zero-shot CoT prompting  \textit{“Let’s think step by step”} \cite{kojima2022large}.}
  \label{fig:SLI-app-sample}
\end{figure}

Even though benchmarks for general abilities and some professional fields are widely used, there is still a scarcity of evaluation datasets on SLI and SLI for language learning. We introduce a new dataset \textbf{SLIQ-LL}\footnote{Available in Github:\href{https://github.com/vocaliodmiku/SLI-LL}{https://github.com/vocaliodmiku/SLI-LL}} (\textbf{S}poken \textbf{L}anguage \textbf{I}ntelligence \textbf{Q}uestions for \textbf{L}anguage \textbf{L}earning) that covers topics of \texttt{phonetics, phonology, and second language acquisition}, which are frequently addressed in language learning education. The dataset consists of two parts\footnote{Some concepts that are crucial for application are placed in the Application Questions subset rather than in the Knowledge \& Concept subset. They are usually of lower difficulty in the Knowledge \& Concept subset.}: %
\begin{itemize}
    \item \textbf{Knowledge \& Concept}: A set of concepts that are mainly designed to test the large models' knowledge of spoken language, such as \textit{"what is language transfer"} and \textit{"how many classifications of consonants are there for the manner of articulation?"} It is mainly sampled from  \cite{ladefoged2014course}. 
    \item \textbf{Application Questions}: To address the ever-changing personalized problems, it is necessary to utilize knowledge of phonetics or linguistics to make complex reasoning. For example, different contexts call for different appropriate stress patterns, which is fundamental to automatic personalized language learning system. We designed this section of data by considering aspects such as \texttt{pronunciation, pauses, stress}, and \texttt{intonation}. An example about \texttt{pronunciation} is shown in the Figure~\ref{fig:SLI-app-sample}.
\end{itemize}

\begin{figure}[h]
  \centering
  \includegraphics[width=\linewidth]{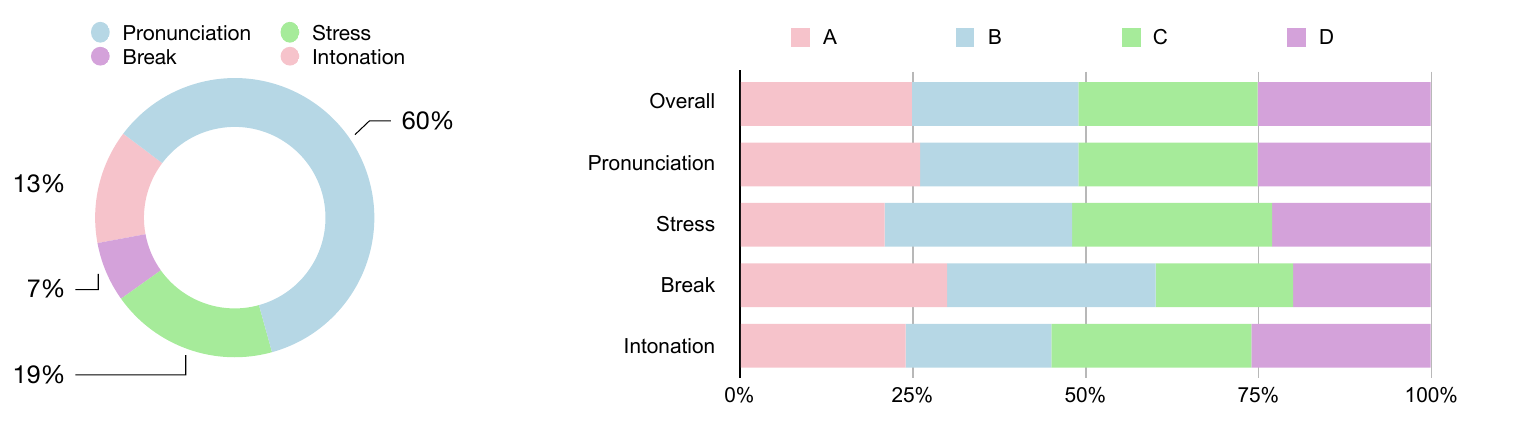}
  \caption{\textbf{Left}: Distribution of problem types in the \textbf{Application Questions} subset. We labeled each question based on the problem and all corresponding options. Each question may have one or multiple types. \textbf{Right}: Distribution of four answer options. \textbf{Overall} refers to the entire dataset. We report the distribution of answer options for different problem types in the \textbf{Application Questions} subset.}
  \label{fig:SLI-dataset-type-dist}
\end{figure}

We collected and designed a total of 445 questions, including 144 Knowledge \& Concept and 301 Application Questions. Figure~\ref{fig:SLI-dataset-type-dist} (left) shows the type distribution of Application Questions, with \texttt{pronunciation} accounting for 60\% \footnote{A few decades ago Gimson and Ramsaran \cite{roach_1971} considered that achievement in acquiring a language comprised nearly 100\% understanding of its pronunciation, $50\sim90\%$ of its grammar, and only 1\% of its vocabulary.}. In the context of language learning, during our initial preparations, we found that intermediate-level learners need to focus more on fundamental pronunciation issues, as opposed to suprasegmental features like stress, break, and intonation. To present this data, we have formulated multiple-choice questions, each with only one correct answer. We uniformly assigned their answers, and Figure~\ref{fig:SLI-dataset-type-dist} (right) shows the statistics.

\section{Method}
This paper delves into various prompt engineering methods for spoken language question-answering, with the aim of gaining a comprehensive understanding of the genuine performance of diverse models. The prompt templates are summarized in Table~\ref{tab:prompts-table}.
\begin{table}[h]
\caption{Prompt templates. In the table below, we use blue and square brackets {\color{niceblue}\texttt{[provided data]}} to represent the data provided to the model, and red and angle brackets {\color{nicered}\texttt{<completions>}} to represent the parts that the model needs to generate. The symbol $\emptyset$ represents an empty string. Self-consistency was not included in the table.}
\label{tab:prompts-table}
\begin{center}
\resizebox{\columnwidth}{!}{%
\begin{tabular}{l|l|l}
\toprule 
Task Description / System & \multicolumn{2}{l}{You are an expert in phonetics, English phonology and second language acquisition.} \\
 & \multicolumn{2}{l}{Here is a multiple-choice question for you to answer correctly.} \\
\toprule 
{} & \bf Zero-shot & \bf Few-shot \\
\addlinespace[0.1cm]
Shot &  $\emptyset$  & Question: {\color{niceblue}\texttt{[Question]}} \\
     &               & Answer:   {\color{niceblue}\texttt{[answer]}} \\
     &               & ... \\
Question & Question: {\color{niceblue}\texttt{[Question]}} & Question: {\color{niceblue}\texttt{[Question]}} \\
Answer & Answer:   {\color{nicered}\texttt{<answer>}} & Answer:   {\color{nicered}\texttt{<answer>}} \\
\midrule 

{} & \bf Zero-shot CoT & \bf Few-shot CoT \\
\addlinespace[0.1cm]
Shot &  $\emptyset$  & Question: {\color{niceblue}\texttt{[Question]}} \\
     &               & Answer: Let's think step by step. {\color{niceblue}\texttt{[CoT]}} \\
     &               & Therefore, the answer is {\color{niceblue}\texttt{[answer]}} \\
     &               & ... \\
Question & Question: {\color{niceblue}\texttt{[Question]}} & Question: {\color{niceblue}\texttt{[Question]}} \\
CoT & Answer: Let's think step by step. {\color{nicered}\texttt{<CoT>}} & Answer: Let's think step by step. {\color{nicered}\texttt{<CoT>}} \\
Answer & Therefore, the answer is {\color{nicered}\texttt{<answer>}} & Therefore, the answer is {\color{nicered}\texttt{<answer>}} \\
\midrule 

{} & \bf In-domain exampler & \bf Tool Augmentation \\
\addlinespace[0.1cm]
Question &  Question: {\color{niceblue}\texttt{[Question about $T$\textsuperscript{$\star$}]}}  & Question: {\color{niceblue}\texttt{[Question]}} \\
&  Answer: Let's think step by step. {\color{niceblue}\texttt{[CoT]}}  & Thought: {\color{nicered}\texttt{<CoT>}} + I need to use some tools to find the answer. \\
&  Therefore, the answer is {\color{niceblue}\texttt{[answer]}}  & Action: \{"action": "{\color{nicered}\texttt{<tool>}}", "input": "{\color{nicered}\texttt{<input>}}"\} \\
&  ... & Observation: The {\color{nicered}\texttt{<tool>}} contains detailed information about the {\color{nicered}\texttt{<input>}}. \\
& Question: {\color{niceblue}\texttt{[Question about $T$]}} & ...\\
CoT & Answer: Let's think step by step. {\color{nicered}\texttt{<CoT>}} & Thought: {\color{nicered}\texttt{<CoT>}} \\
Answer & Therefore, the answer is {\color{nicered}\texttt{<answer>}} & Therefore, the answer is {\color{nicered}\texttt{<answer>}} \\
\bottomrule
\addlinespace[0.1cm]
\multicolumn{3}{l}{\footnotesize{
\textsuperscript{$\star$}$T \in \{\mathrm{Pronunciation}, \mathrm{Stress}, \mathrm{Break}, \mathrm{Intonation}\}$.
}}\\
\end{tabular}%
}
\end{center}
\end{table}·

\paragraph{Zero-shot} We adopt zero-shot as the benchmark for our experiment. This approach involves simply posing a question and requesting an answer, which is the most straightforward and widely used way to leverage LLMs. However, a foundation model that has not been fine-tuned on any task or instruction fine-tuned, particularly one with a limited parameter size, may not generate meaningful responses to user inputs. Nonetheless, as our test data are presented in the form of multiple-choice questions, which aligns with many LLM benchmarks \cite{hendrycks2020measuring,lai2017race,lin2021truthfulqa}, this technique is relatively user-friendly and practical \cite{robinson2022leveraging}.

\paragraph{Few-shot} We insert multiple examples in both the task description and the question-answer pairs, each of which is structured as a zero-shot setting. Additionally, the CoT few-shot configuration includes an explanation component for each example.

\paragraph{CoT} We further incorporate reasoning chains in the prompts, whereby a \textit{"Let's think step by step"} prompt is added to the end of the prompt in the zero-shot setting. This is followed by a statement to draw the conclusion: \textit{"Therefore, the answer is }{\color{nicered}\textit{<completions>}}". 

\paragraph{In Domain exampler}
When selecting examplar samples for each question, we match them with more relevant examples based on the type of question. For example, in the Application Question subset, if the current question type is related to \texttt{pronunciation}, then the question type used to construct the current examples will also be related to \texttt{pronunciation}. 
This configuration is based on few-shot CoT.

\paragraph{Self-Consistency} CoT prompting offers more options to deduce the answer to a given question, with a primary focus on generating multiple lines of reasoning and striving to find a consensus among the resulting answers (such as selecting the most consistent answer through voting among these paths) \cite{wang2022self,imani2023mathprompter}. Self-consistency can even improve certain tasks where CoT prompting traditionally falls short of standard prompting \cite{wang2022self}. The word in \cite{wang2022rationale} showed that diverse reasoning paths are the critical factor in improving CoT reasoning performance. The integration of self-consistency into CoT prompting can readily enhance performance without requiring additional training. Here, we utilize multiple answers generated by CoT to obtain self-consistency results through majority voting.

\paragraph{Tool Augmentation} Although LLMs can memorize some of the knowledge ingrained in the training data, they may still struggle to utilize this knowledge efficiently during inference. To enhance the performance of language models, a branch of research such as the exploration of external tools and retrieval of relevant information from a knowledge base has be considered \cite{nakano2021webgpt,schick2023toolformer,gao2023pal,lewis2020retrieval,borgeaud2022improving}. We utilize \textit{Google} and \textit{Wikipedia} as external tools to assist in answering the question. We use \textit{langchain} toolkit\footnote{https://github.com/langchain-ai/langchain} to conduct the experiment and this prompt is only tested in the zeor-shot learning setting.

\section{Experiments} 
\paragraph{Models} We report results across 20 unique LLMs including 4 model families (GPT-3.5 \& GPT-4 \cite{openai2023gpt4}\footnote{The API we used has a suffix of "0613", i.e. \texttt{GPT-4-0613}.}, LLaMA 1 \& LLaMA 2 \cite{touvron2023llama,touvron2023llama2}, FLAN-T5 \& UL2 \cite{tay2022unifying,chung2022scaling}, and Pythia \cite{biderman2023pythia}) on SLI-LLM dataset. See Appendix~\ref{apdx:overall-llm} and Appendix~\ref{apdx:overall-results} for more details.


\paragraph{Implementation details} We conduct our experiment in two rounds. In the first round, we perform a comprehensive performance review on a large scale. We consider two prompting strategies, including the direct and  CoT prompting, under both zero-shot and few-shot (3-shot) learning paradigms. In the second round, we carefully analyze representative models with advanced prompt methods. All reasoning is done using the parameters \verb|{max_new_tokens: 512, temperature: 0, top_p: 1}| to obtain the most deterministic results. We write five develop examples with thought chains for Concept and Aapplications Questions, respectively, and unless specified, few-shot examples are sampled from these five questions. We manually review certain answers that cannot be identified by regular expressions. This process also demonstrates the model's ability to follow instructions.

\section{Results and Analysis}
\subsection{Zero-Shot \& Few-Shot Benchmark}

\begin{figure}[h]
  \centering
  \includegraphics[width=\linewidth]{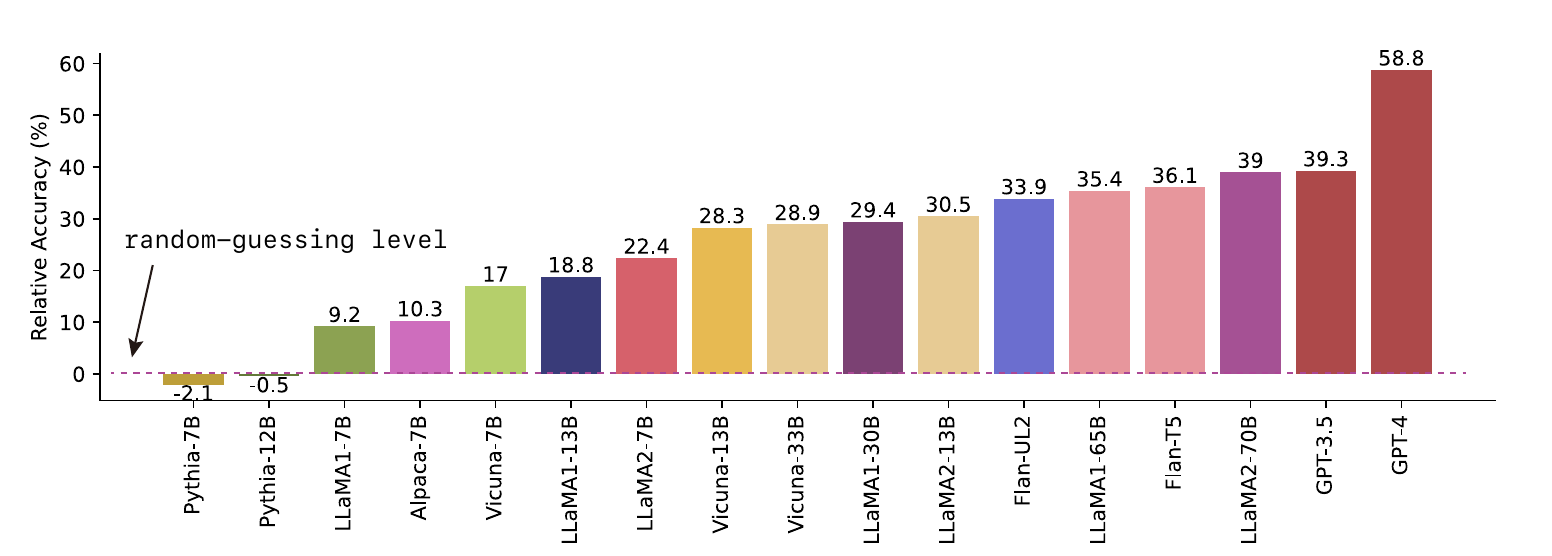}
  \caption{Overall performance on the SLIQ-LL dataset. We report the best results of each model among four different prompting methods. The results in the figure are the raw results minus 25\% (random selection level).}
  \label{fig:base_analyse}
\end{figure}

We present the full results of the first round in Table~\ref{tab:overall-res} in Appendix~\ref{apdx:overall-results}. 
Figure~\ref{fig:base_analyse} demonstrates the model's ability by showing the best performance among four prompting methods. 
It should be noted that the results in the figure are the raw results minus 25\% (random selection level).
Overall, models with larger parameter sizes show better overall performance. Despite this, Pythia struggles with generating reasonable responses. The performance of the best prompting strategy is still below the level of random guessing. In contrast, GPT-4 displays exceptional performance with a significant advantage. Surprisingly, Flan-T5 has a comparable performance to models with several times their parameter size and the 70B version of LLaMA2 also achieves a competitive performance to GPT-3.5. The current best performance of open-source models stands at 61.6\%, highlighting the potential for further optimization. To gain a more precise understanding of these models' performance, we conduct a more in-depth analysis of various aspects, as outlined below.

\begin{figure}[h]
  \centering
  \includegraphics[width=0.95\linewidth]{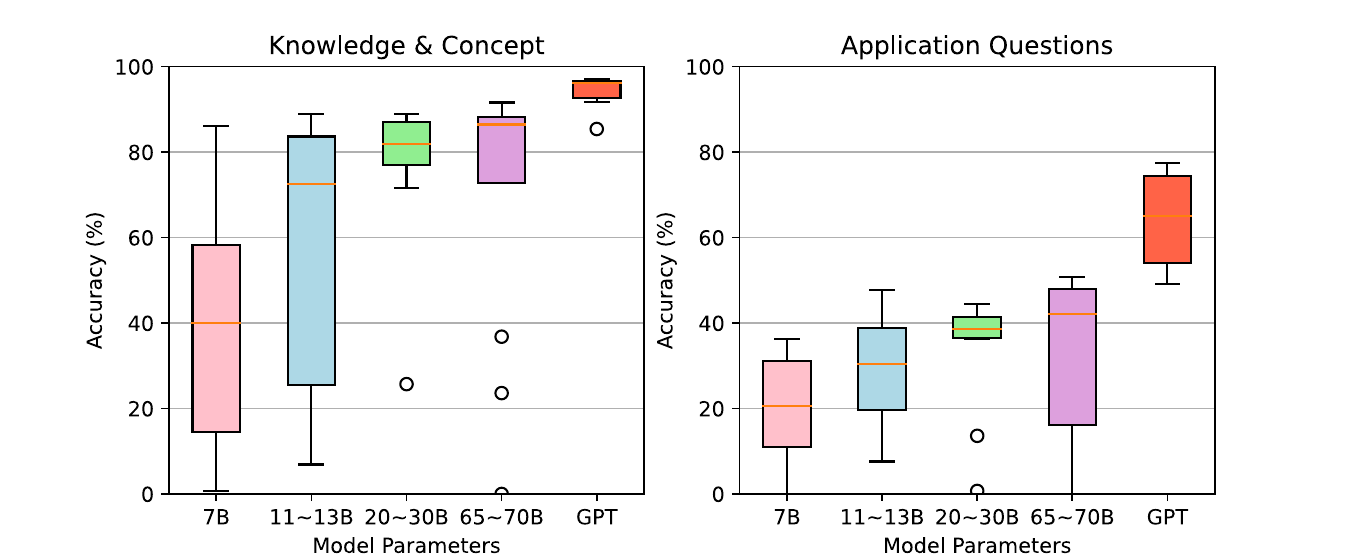}
  \caption{The distribution of performance on two subsets across different model sizes.}
  \label{fig:SLI-Subset-Perf}
\end{figure}

\paragraph{Models with more parameters tend to have better performance and stability in subsets} As we examine the two data subsets more closely (as depicted in Figure~\ref{fig:SLI-Subset-Perf}), we can draw a similar conclusion as the above. Notably, regarding Knowledge \& Concept, models with a parameter size exceeding 20B display reduced performance variation, signifying heightened stability in their performance. 

\paragraph{LLM excels in concept memorization but has weaker ability in applying knowledge for reasoning}Even on a relatively small model (7B), the accuracy of concept memorization can reach nearly 80\%, and a model with a size of around 11B can achieve the level of GPT-3.5. At around 20B, LLMs reach performance saturation in Knowledge \& Concepts. However, for reasoning Application Questions, even the most powerful LLaMA2 models with sizes of 70B and the GPT series models still perform relatively low accuracy (42.6\%, 64.3\%).

\begin{figure}[h]
  \centering
  \includegraphics[width=0.6\linewidth]{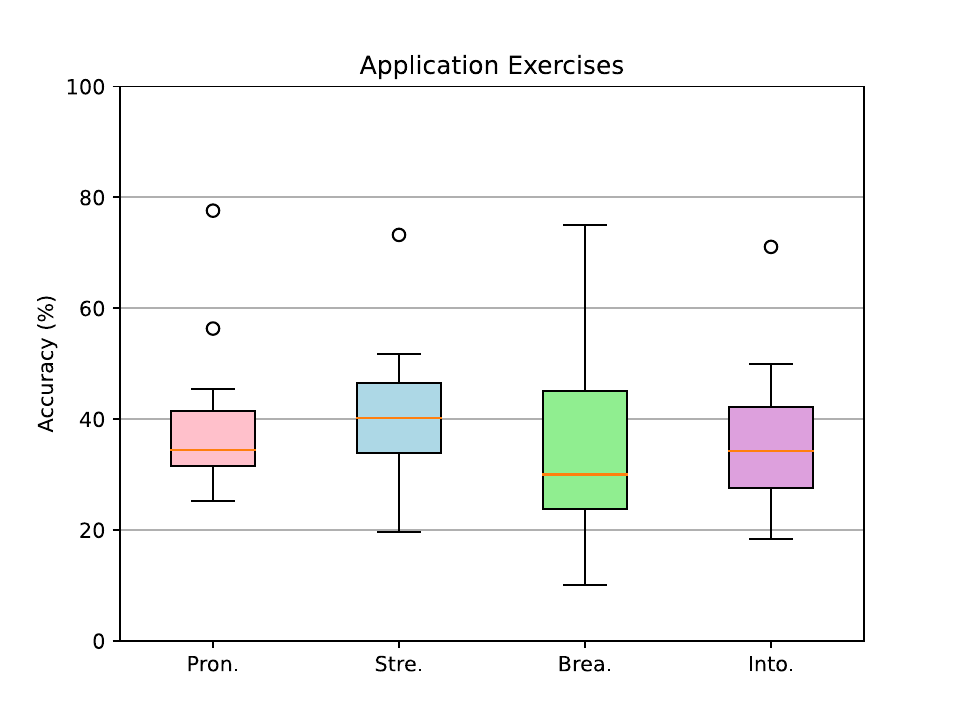}
  \caption{The accuracy distribution across different question types in the Application Questions subset.}
  \label{fig:SLI-type-Perf}
\end{figure}

\paragraph{Knowledge preference} We analyzed whether these models exhibit any specific preferences for certain types of knowledge. Figure~\ref{fig:SLI-type-Perf} demonstrates that these models display no significant differences in terms of accuracy among different types of questions. Furthermore, their performance on the question type \texttt{break} is near to random guessing.

\begin{table}[h]
  \caption{Frequencies of predictions and labels. We highlight labels that are under estimated using the color {\color{niceblue} blue $\blacktriangledown$} and over estimated using the color {\color{nicered} red $\blacktriangleup$} ($\pm$ 20\% of the label frequency). Using the $\chi^2$ test, we report the p-value for the null hypothesis "\textit{the predictive distribution equals the empirical one}".}
  \label{tab:SLI-bias}
  \begin{center}
  \vspace{-1.0em}
  \resizebox{0.85\linewidth}{!}{%
  \begin{tabular}{lllllll}
\toprule
\bf Model & \bf A & \bf B & \bf C & \bf D & \bf Acc. & \bf $p$-value \\
\midrule
GPT-4 & 77 & 84 & 69 & 71 & 74.6\% & $5 \cdot 10 ^{-1}$ \\
GPT-3.5-turbo & 76 & 91 & 70 & 64 & 53.1\% & $2 \cdot 10^{-1}$ \\
LLaMA2-70B-chat & 64 & 87  & {\color{nicered} 100$\blacktriangleup$} & {\color{niceblue} 50$\blacktriangledown$} & 44.3\% & $2 \cdot 10 ^{-3}$ \\
Flan-UL2-20B & {\color{nicered} 93$\blacktriangleup$} & {\color{niceblue} 56$\blacktriangledown$}  & 70 & 82 & 41.6\% & $4 \cdot 10 ^{-3}$ \\
Flan-T5-XXL-11B & {\color{nicered} 93$\blacktriangleup$} & 67 & 80 & 61 & 44.8\% & $6 \cdot 10 ^{-2}$ \\
\hline \addlinespace[0.05cm]

\bf data & \bf 74 &\bf  76 & \bf 80 & \bf 71 & \\
\bottomrule
\addlinespace[0.05cm]
\multicolumn{7}{l}{\small{Averaged using $\{0,3,5,8,10,12,15,20,25,30\}$ shots and $1\sim9$ shots CoT results.
}} \\
\multicolumn{7}{l}{\small{Part of the experimental data (i.e., CoT) comes from Section~\ref{sec:fscot}.}}
\end{tabular}
  }
  \end{center}
\end{table}

\paragraph{Answer Bias} We selected several models and conducted corresponding distribution analysis of generated answers in Table~\ref{tab:SLI-bias}. It can be observed that, apart from GPT-3.5 and GPT-4, the other models exhibit apparent answer bias.

\subsection{Advanced Prompt Analysis}
In the large-scale experiment, we utilized an empirical value of $k=3$ as the number of shots and have designed them for two subsets. In this section, we explored additional possibilities.

\begin{figure}[h]
  \centering
  \includegraphics[width=1\linewidth]{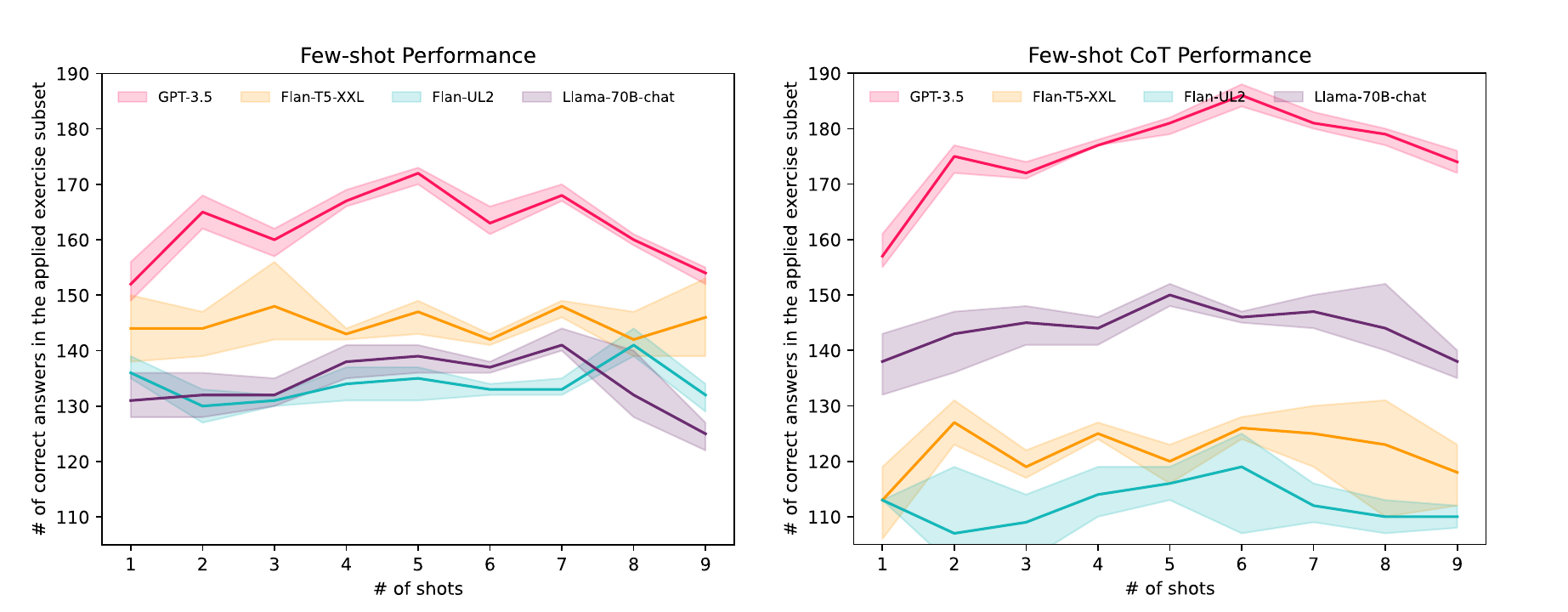}
  \caption{Number of correctly answered questions (total 301) on SLIQ-LL (Application Questions subset) using $k=1...9$ shots. The develop examples used in the previous experiments were manually written, while the current ones are sampled from within dataset. The thought chain is generated by GPT-4 with correctly generated answers (the thought chain is not guaranteed to be correct). As no specific conclusions can be drawn, we are not reporting the performance of few-shot direct with $k > 9$.}
  \label{fig:SLI-fs-Perf}
\end{figure}
\paragraph{Few-shots \& CoT}\label{sec:fscot} Figure~\ref{fig:SLI-fs-Perf} presents the results of direct few-shot and CoT few-shot using different numbers of examples. Based on the results, we can see that increasing the number of examples can improve performance to a limited extent and that increasing the examples of the reasoning chain will have a more significant and stable effect on models above 70B (LLaMA2-chat, GPT-4). However, for smaller models, these prompts may already exceed their capabilities, resulting in degradation of performance.

\begin{figure}[h]
  \centering
  \includegraphics[width=0.7\linewidth]{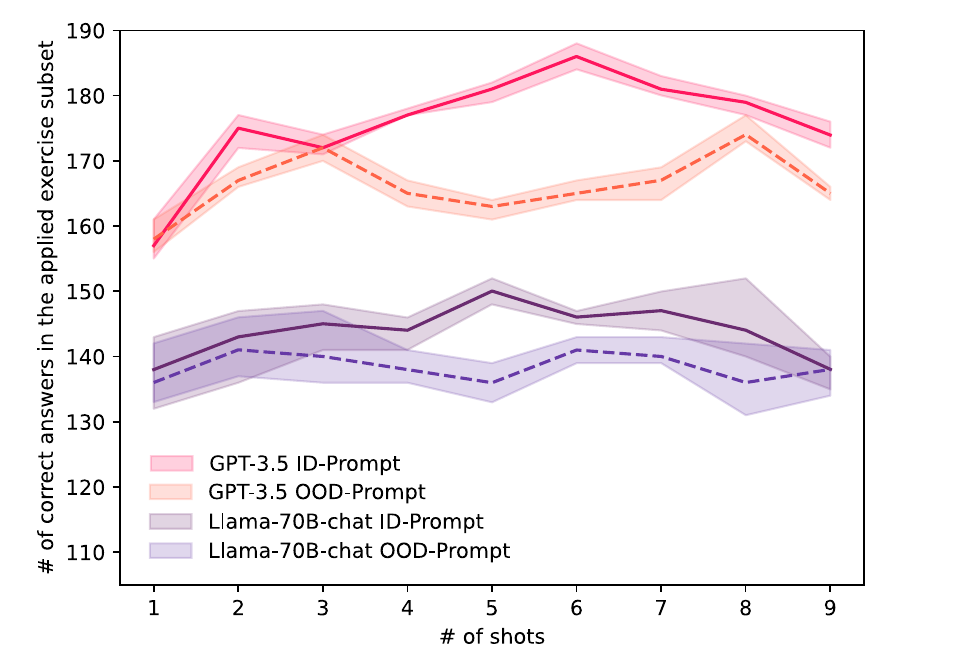}
  \caption{Few-shot CoT performance: In-Domain prompt v.s. Out-Of-Domain (Commonsense) prompt. We are creating examples that are customized for each question type. For instance, when the current question concerns \texttt{grammar}, the provided examples will be \texttt{grammar}-related as well. This is referred to as \textbf{In-Domain}. We randomly picked one \textbf{Out-Of-Domain} (OOD) sample of 20 disciplines from the MMLU dataset with CoT \cite{fu2023chainofthought} and conducted manual checks on these examples.}
  \label{fig:SLI-Model-Size-Perf}
\end{figure}

\paragraph{In-Domain Prompt v.s. Out Of Domain Prompt} We used prompts from different domains for two models capable of responding to CoT. In most cases, most examples are not carefully selected or designed. We used domain-specific prompts for different types of questions. In Figure~\ref{fig:SLI-Model-Size-Perf}, our approach has shown significant advantages compared to more common examples. In the process of increasing examples, the model not only learned how to answer multiple-choice questions, but also gained some insights.

\begin{table}[h]
\caption{Compare self-consistency with CoT on the GPT-3.5 and LLaMA2-70B-CHAT models.}
  \label{tab:SLI-SC}
  \begin{center}
  \begin{tabular}{lcc}
\toprule
\bf Model & \bf CoT & \bf Self-Consistency \\
\midrule
GPT-3.5 & 60.1\pmx{1.5} & \bf 64.4 \\
LLaMA2-70B-chat & 48.6\pmx{1.2} & 48.2 \\
\bottomrule
\addlinespace[0.05cm]
\multicolumn{3}{l}{\small{
Using top 5 few-shots CoT results.
}}
\end{tabular}
  \end{center}
\end{table}

\paragraph{Self-Consistency} Although the solution to these phonological problems does not have as many reasoning paths as mathematical reasoning questions, we found that self-consistency can improve performance on the GPT-3.5 model (as shown in Table~\ref{tab:SLI-SC}). However, for LLaMA2-70b-chat, its occasional cleverness can be offset by multiple generated errors. Appendix~\ref{apdx:sc-sample} lists examples where GPT-3.5 successfully corrected errors while LLaMA2-70b-chat failed unfer Self-Consistency.

\begin{table}
  \caption{The number of API calls made using GPT-3.5.}
  \label{tab:SLI-api}
  \begin{center}
  \begin{tabular}{lc}
\toprule
\bf Tool & \bf Request Times \\
\midrule
Google & 351 \\
Wikipedia &  131 \\
\bottomrule
\addlinespace[0.05cm]
\end{tabular} 
  \end{center}
\end{table}

\begin{table}
  \caption{The performance of tools augmented GPT-3.5. We use Google and Wikipedia. \textbf{Explicit Reject} represents the model's explicit rejection of the question, such as: \textit{'Based on the available information, the answer cannot be determined.'} \textbf{True Reject} means that the model's rejection avoids generating incorrect answers in zero-shot generation."}
  \label{tab:SLI-rej}
  \begin{center}
  \begin{tabular}{lccc}
\toprule
\bf Method & \bf Acc. & \bf Explicit Reject & \bf True Reject  \\
\midrule
Zero-Shot & 49.1 & 1 & - \\
Tools Aug. &  49.1 & 14 & 6\\
\bottomrule
\addlinespace[0.05cm]
\end{tabular} 
  \end{center}
\end{table}

\paragraph{Augmented Language Models} On the internet, individuals actively share their language learning experiences. By effectively using this external knowledge, LLMs can improve credibility and mitigate hallucination issues. We provided two tools for use by GPT-3.5, but did not get better results (Table~\ref{tab:SLI-api}). One piece of good news is that models using the tools can recognize their limitations and refuse to answer questions they are uncertain about, although this ability still seems relatively limited (Table~\ref{tab:SLI-rej}).

\subsection{Expert evaluation of the Chat interface}
Compared to single-turn Q\&A, people prefer interacting through dialogue interfaces. Here, we conducted a more challenging evaluation in the context of language learning, specifically CAPT. We sampled 20 sentences from Chinese English learners' speech with an average score ranging from 0 to 100. The mispronunciation, context, and prosodic information of each sentence were included. We organized this information as input for a chat model, and then engaged in discussions about improving pronunciation based on this information. This evaluation was designed to assess the ability of the model to analyze and reason a given question using knowledge acquired through phonetics and second language acquisition in a longer contextualized setting. In Appendix~\ref{apdx:capt-examples}, we present examples and the evaluation methodology. We evaluate the model's answers from the following perspectives.
\begin{itemize}[label={}]
  \item \hlc[AA!40]{RATING-A}: The response is both valid and satisfactory, and is relevant to the evaluation prompt.
  \item \hlc[BB!40]{RATING-B}: The response is acceptable, but with minor errors or imperfections.
  \item \hlc[CC!40]{RATING-C}: Although the response is relevant and addresses the instruction, it contains significant errors in its content.
  \item \hlc[DD!40]{RATING-D}: The response is either irrelevant to the evaluation prompt or entirely invalid for current topic.
\end{itemize}
In Table~\ref{tab:SLI-chat-res}, we report the results of expert evaluation. GPT-3.5 achieves high performance and reliability. If we consider rating \hlc[AA!40]{A} and \hlc[BB!40]{B }as acceptable responses, its accuracy reaches 83.4\%, which is nearly identical to its performance on SLIQ-LL. In contrast, despite using relatively fixed prompts, this still posed a challenge for promising model LLaMA2-70B-chat, it only achieved acceptable performance in 54\% of cases.

\begin{table}[h]
  \caption{Distribution(\%) of observed patterns (A, B, C, D) identified among Multi-turn Conversations using 20 CAPT Samples.}
    \label{tab:SLI-chat-res}
    \begin{center}
    \begin{tabular}{lcccc}
\toprule
\bf Model & \color{AA}{\bf A} & \color{BB}{\bf B} & \color{CC}{\bf C} & \color{DD}{\bf D} \\
\midrule
GPT-3.5 & 55.6 & 27.8 & 16.7 & 0.0 \\
LLaMA2-70B-chat & 35.1 & 18.9 & 43.2 & 2.7 \\
\bottomrule
\addlinespace[0.05cm]
\end{tabular}
    \end{center}
  \end{table}

\section{Discussion}
In this paper, we conducted several experiments based on prompt engineering. Our designed extraction of conceptual knowledge posed little challenge for these LLMs. However, the models encountered some difficulties when using this knowledge for inference. For some small models, due to excellent instruction fine-tuning, their responses were almost always valid (meaning they could generate responses similar to \textit{"So the answer is A"}). The more valid responses, the more likely the correct answers were. However, for many small or even relatively large models, their performance on the application datasets was worse than random guessing (see Appendix~\ref{apdx:overall-results}). This was largely because they could not generate valid outputs following instructions. Therefore, we declare that the results we reported are actually a comprehensive reflection of both SLI and the models' ability to follow instructions (which is why we reported the best results of the models in different prompt modes in Figure~\ref{fig:base_analyse}).

Some widely proven effective methods combined with the domain-specific examples resulted in significant performance improvements for the models (GPT-3.5, 49.1\% -> 63.1\%; LLaMA2-70B-Chat, 42.2\% -> 48.6\%). However, these performance improvements were limited to larger models. In most cases, appending more examples in direct and CoT scenarios did not seem to yield stable performance improvements. Given that the number of consumed tokens is increasing, we consider $3\sim5$ shots a reasonable choice.


We conduct an investigation using external tools. After leveraging knowledge from Google and Wikipedia, GPT-3.5 successfully achieves performance comparable to zero-shot scenarios. This outcome suggests that these questions are not easily resolved through internet searches. A reliable alternative approach is to establish a dedicated knowledge repository as an additional source of information.

During our evaluation in dialogue mode, we have observed that GPT-3.5 demonstrates a high level of usability. It maintains a strong focus on the subject and content of the conversation, showing minimal tendency to veer off-topic as the dialogue progresses (see Sample~\ref{tab:slill-sample-capt-GPT-3.5}). Its reasoning abilities remain consistent, comparable to its performance in single-turn tests. In contrast, LLaMA2-70b-chat often becomes perplexed by its own generated responses, digressing into self-referential narratives and forgetting to prioritize the user's input (see Sample~\ref{tab:slill-sample-capt-llama-f}). The interactive nature of dialogue-based language interaction continues to be a captivating approach, especially in the field of language learning, consistently attracting the interest of researchers and developers in the industry.

\section{Conclusion}

We explored zero-shot, few-shot, direct, and CoT prompts to phonology-related questions answering. These models all have strong conceptual knowledge and can achieve high accuracy with simple zero-shot and few-shot learning. For practical questions reasoning, we achieved significant performance improvements compared to the zero-shot baseline (GPT-3.5, 49.1\% -> 63.1\%; LLaMA2-70B-Chat, 42.2\% -> 48.6\%). However, the strongest GPT-4 achieved 77.4\% accuracy. This means there is significant room for improvement in their performance in real-world scenarios. These performances highlight the impressive Spoken Language Intelligence exhibited by LLMs, and Chatbots based on large language models possess significant potential to enhance conversational spoken language learning.

\begin{limi}
As described in the introduction~\ref{para:SLI}, it is important to conduct \textbf{multimodal evaluation for SLI}. However, the evaluation of LLMs presented in this paper only utilizes textual data and does not investigate the performance of multimodal models, including those based on speech or images.  \cite{zhang2023m3exam} constructed a benchmark dataset to evaluate Multimodal Large Language Models (MLLM). However, the evaluation benchmark, particularly for the speech modality, is still quite limited. In theory, the measurement of SLI should place greater emphasis on acoustic features. For instance, when presented with a speech segment from a conversation, it is important to determine if a LLM equipped with speech input can accurately identify which vowel is being spoken at any given moment, along with associated parameters such as the sampling rate, duration, F1, F2, and other relevant downstream information. To differentiate from ASR, prompts like \textit{"Is this a high or low vowel?"}, \textit{"Is there background piano music?"} or \textit{"How many speakers are present?"} might be utilized. 

Furthermore, in language learning settings, it is important to consider performance in \textbf{multilingual scenarios}. For instance, Chinese English learners may prefer feedback presented in their native language rather than English. Therefore, it is crucial to pay attention to such factors and cater to the needs of learners to ensure effective language acquisition.

\end{limi}

\begin{ack}
  This work is partly supported by the Fundamental Research Funds for the Central Universities (No. 2023RC13).
\end{ack}

\bibliographystyle{unsrt}
\bibliography{SLI_2023}

\clearpage
\newpage
\appendix
\section{Statistics of LLMs we used}\label{apdx:overall-llm}
\begin{table}[h]
    \caption{Statistics of large language models used in this work, including the capacity evaluation, pre-training data scale (either in the number of tokens or storage size) and hardware resource costs. The term ``Adaptation'' refers to whether the model has been with subsequent fine-tuning: IT denotes instruction tuning and RLHF denotes reinforcement learning with human feedback. ``Evaluation'' indicates whether the model has been evaluated with corresponding abilities in their original paper: ICL representing in-context learning and CoT representing chain-of-thought. Most of the model checkpoints can be publicly accessible while except GPT-3.5 and GPT-4. GPT-3.5 is an upgraded version of GPT-3 with RLHF and \texttt{GPT-3.5-turbo} belongs GPT-3.5 series, which is the interface to invoke Chat-GPT.}
    \label{tab:resource_model}
  \centering
  \resizebox{\linewidth}{!}{
  \begin{tabular}{lccccccccccc}
    \toprule
    &   & \multicolumn{1}{c}{}  & & \multicolumn{2}{c}{\textbf{Adaptation}} &   &    && \multicolumn{2}{c}{\textbf{Evaluation}}   \\
    \multirow{-2}{*}{\textbf{Model}} & \multirow{-2}{*}{\textbf{\begin{tabular}[c]{@{}c@{}}Release\\ Time\end{tabular}}} & \multicolumn{1}{c}{\multirow{-2}{*}{\textbf{\begin{tabular}[c]{@{}c@{}}Size\\ (B)\end{tabular}}}} & \multirow{-2}{*}{\textbf{\begin{tabular}[c]{@{}c@{}}Base\\ Model\end{tabular}}} & \textbf{IT}   & \textbf{RLHF} & \multirow{-2}{*}{\textbf{\begin{tabular}[c]{@{}c@{}}Pre-train\\ Data Scale\end{tabular}}} & \multirow{-2}{*}{\textbf{\begin{tabular}[c]{@{}c@{}}Hardware\\ (GPUs / TPUs)\end{tabular}}} & \multirow{-2}{*}{\textbf{\begin{tabular}[c]{@{}c@{}}Training/FT\\ Time\end{tabular}}} & \textbf{ICL} & \textbf{CoT} \\
    \midrule
      LLaMA~\cite{touvron2023llama}    & Feb-2023    & $7\sim 65$    & -   & - & - & 1.4T tokens   & 2048 80G A100 & 21 d   & $\checkmark$ & -    \\
      Vicuna~\cite{zheng2023judging}    & Mar-2023    & $7\sim 33$    & LLaMA   & - & - & -  & 8 80G A100 & -   & $\checkmark$ &$\checkmark$    \\
      Alpaca~\cite{alpaca}    & Mar-2023    & 7    & LLaMA   & $\checkmark$ & - & - & 4 80G A100 & 3 hrs   & -  & -    \\
      Flan-T5~\cite{chung2022scaling}  & Oct-2022    & 11 (XXL)    & T5   & $\checkmark$ & - & -  & -  & -  & $\checkmark$ & $\checkmark$    \\
      Flan-UL2~\cite{tay2022unifying}  & Mar-2023    & 20    & UL2   & $\checkmark$ & - & - & - & -  & $\checkmark$ & $\checkmark$    \\
      Pythia~\cite{biderman2023pythia}  & Apr-2023    & 12   & - & -  & - & 300B tokens & 256 40G A100   & -  & $\checkmark$ & - \\
      LLaMA2~\cite{touvron2023llama}    & Jul-2023    & $7\sim 70$  & - & $\checkmark$ & $\checkmark$ & 2.0T tokens   & 2000 80G A100 & -  & $\checkmark$ & $\checkmark$    \\
    \midrule
    \midrule
      GPT-3~\cite{brown2020language}    & May-2020    & 175   & -   & - & - & {300B tokens} & -  & -  & $\checkmark$ & -       \\
      GPT-4~\cite{openai2023gpt4}    & Mar-2023    & - & -   & $\checkmark$  & $\checkmark$  & - & - & -  & $\checkmark$ & $\checkmark$  \\
    \bottomrule
    \addlinespace[0.07cm]
    \multicolumn{9}{l}{This table is mainly adapted from \cite{zhao2023survey}}
\end{tabular}
  }
\end{table}

\section{Summary of the results}\label{apdx:overall-results}

In Table~\ref{tab:overall-res}, We are presenting the results of large-scale evaluation using the SLIQ-LL dataset. We are using horizontal lines to categorize the different model types based on their size and base model. We report the zero-shot, few-shot($k=3$), and CoT results of these models, and separately providing the results for the subset. The performance gap between this experiment and the zero-shot experiment is represented by $\Delta$. Some marks are utilized to signify that the model's evaluation has failed for different reasons, leading to subpar performance. Mark \#1 represents that he achieved the best performance among models of the same level.

\begin{itemize}[label={}]
\item \textsuperscript{$\star$} Unable to generate an effective response.  
\item \textsuperscript{$\dagger$} The response lists more examples of problems, most of which do not address the original question. 
\item \textsuperscript{$\ddagger$} The response repeats prompts question without providing new information. 
\end{itemize}

\section{Self-Consistency CoT Samples}\label{apdx:sc-sample}
Allowing the model to utilize different reasoning paths is actually a more comprehensive evaluation method, as it reflects the sampling of answers generated by the model for the given problem. If many paths are incorrect, it indicates that the model is unlikely to perform correct reasoning for the problem. We conduct a thorough analysis of the samples in the self-consistency experiment. Similar to  \cite{lievin2022can}, we considered three general skills that we expect are required to be mastered to answer phonological questions: (i) performing non-trivial reasoning steps, (ii) recalling knowledge that is not provided in the context and (iii) the ability to comprehend the question and the context. Based on the three skills, three success patterns (A, B, C) and three failure patterns (D, E, F) are defined.
\begin{itemize}
  \item Table~\ref{tab:sli-sample-sc-GPT-3.5} illustrates how GPT-3.5 recover from occasional errors by utilizing various chains of reasoning.
  \item Table~\ref{tab:sli-sample-sc-llama1} highlights that LLaMA2 is unable to perform correct reasoning due to the generation of excessive erroneous paths.
\end{itemize}

\section{CAPT examples and Promt templates for Chat Interface}\label{apdx:capt-examples}
In Table~\ref{tab:sli-sample-chat-capt}, we present a complete dialogue flow. Firstly, we give the model a system prompt where the data structure of the CAPT results is explained. Then, the example is input as the current object of analysis and a series of questions are posed. We present a multi-turn conversaion example of GPT-3.5 in Table~\ref{tab:slill-sample-capt-GPT-3.5}, an example of LLaMA2-70B-chat in Table~\ref{tab:slill-sample-capt-llama} and a failed example of LLaMA2-70B-chat in Table~\ref{tab:slill-sample-capt-llama-f}. The annotated ranking label for each response is indicated before the answer.

\paragraph{Additional information about the CAPT data} The age range of the speakers is between 15 and 30 years old, and the length of the reference text is between 5 and 15 words. We use the Kaldi toolkit \cite{kaldi} to obtain the GOP \cite{gop} score as the phoneme score through forced alignment. The word score is obtained by averaging the phoneme scores. The ASR system used is Whisper \cite{radford2022robust} Large, and fluency, liaison, break, stress, intonation, and other information are annotated by human experts.

\section{Deepseek R1 performance}\label{apdx:capt-examples}

\begin{figure}[h]
  \centering
  \includegraphics[width=0.5\linewidth]{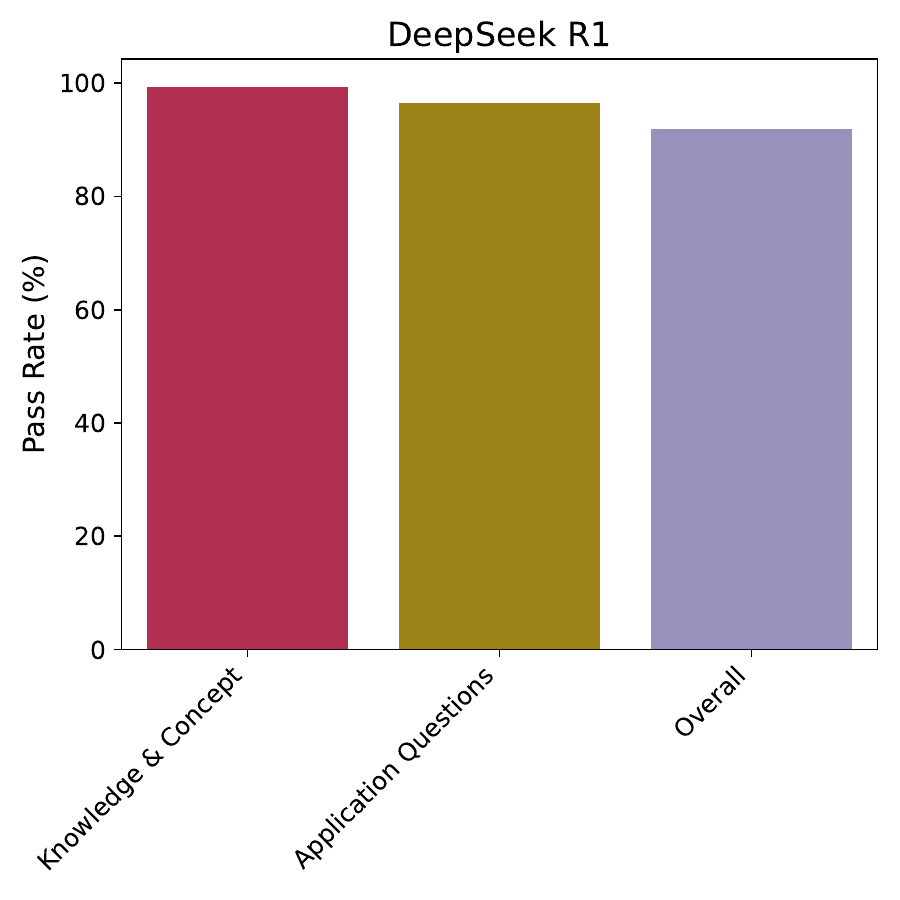}
  \caption{Deepseek R1 performance. We don't use any prompt techniques.}
  \label{fig:DS-Perf}
\end{figure}

\begin{figure}[h]
  \centering
  \includegraphics[width=\linewidth]{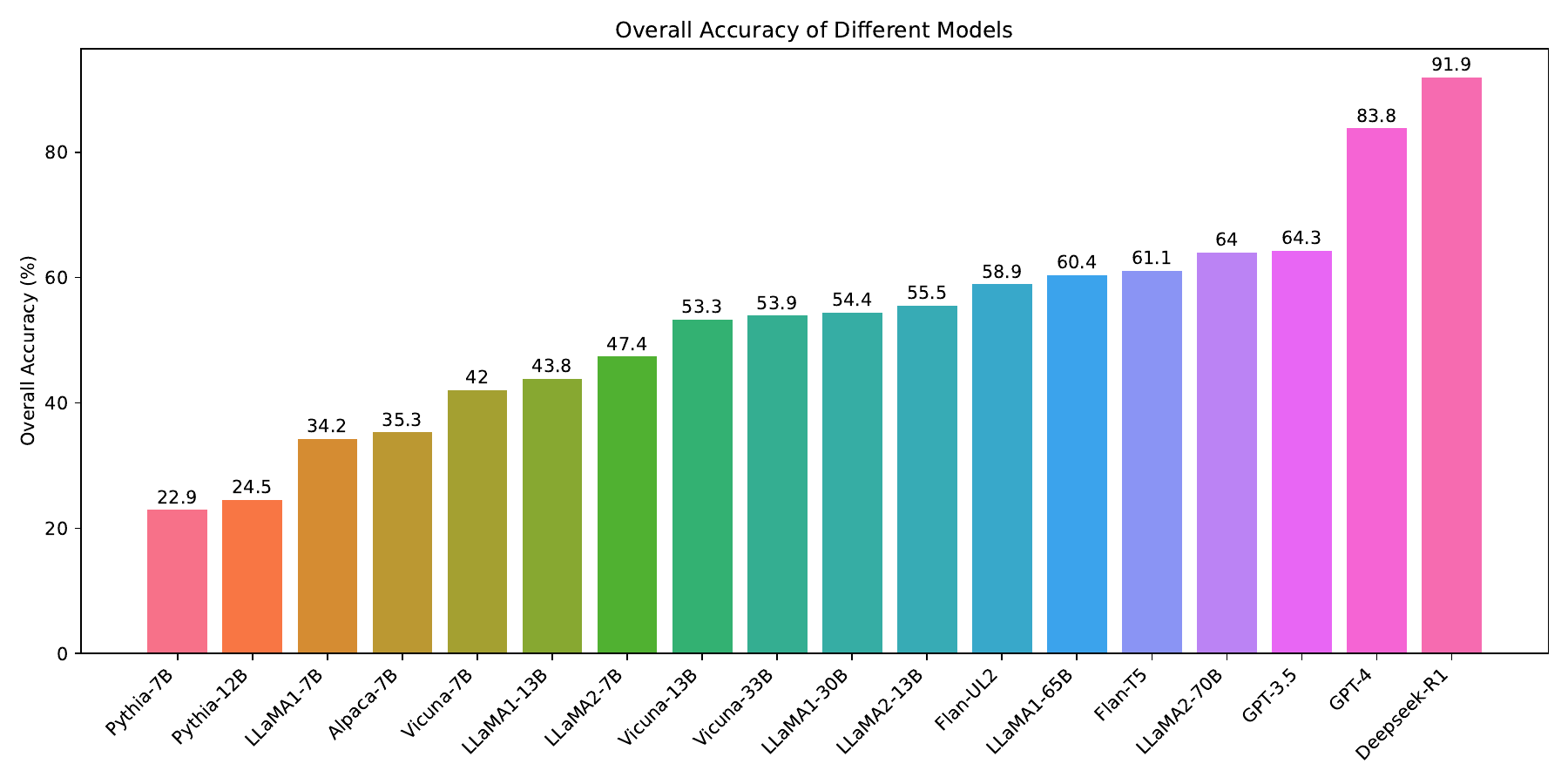}
  \caption{Overall performance on the SLIQ-LL dataset with Deepseek R1 model.}
  \label{fig:DS-Perf-comp}
\end{figure}

In the evaluation of speech and language intelligence, DeepSeek R1\cite{deepseek2023r1} demonstrated exceptional capabilities (Table~\ref{fig:DS-Perf} \& Table~\ref{fig:DS-Perf-comp}). R1 achieved accuracy rates of 99.3\% and 96.5\% in Knowledge \& Concept and Application Questions respectively, with a comprehensive accuracy rate of 91.9\%. This performance significantly surpasses mainstream models such as GPT-3.5 (64.3\%) and GPT-4 (83.8\%), highlighting its technical advantages in the field of language learning. The model, through reinforcement learning (RL)-driven reasoning capabilities, successfully addressed the limitations of traditional large models in applying logic to real-world scenarios.

Additionally, DeepSeek R1's open-source strategy is one of its core highlights. The research team not only released the original models (such as R1-Zero and R1) but also introduced six lightweight versions (ranging from 1.5B to 70B parameters) through knowledge distillation techniques, enabling smaller models to inherit powerful reasoning capabilities. This open approach lowers the technical barrier, allowing communities with limited educational resources to deploy high-performance AI tools at a low cost, thereby promoting educational equity.

\begin{table}
  \caption{\small Evaluation results of various popular models.}
  \label{tab:overall-res}
  \centering
  \resizebox{0.85\linewidth}{!}{
  \begin{tabular}{llcrrrrr}
\toprule
\bf Model & \bf Prompt & \bf Shot & \bf {\color{colorknowledge}Concept} (144) & $\Delta$ &\bf {\color{nicered}Applied Questions}(301) & $\Delta$ & \bf Overall \\
\midrule
LLaMA1-7B & Direct & 0 & {\color{colorknowledge}38.9\% (56)}  &                           & {\color{nicered}0.0\% (0)}\textsuperscript{$\ddagger$}  &                                   & 12.6\%  \\
LLaMA1-7B & Direct & 3 & {\color{colorknowledge}14.6\% (21)}\textsuperscript{$\dagger$}  & {\color{nicered}-24.3\%}  & {\color{nicered}4.0\% (12)}\textsuperscript{$\ddagger$} & {\color{nicegreen}+4.0\%}  & 7.4\%  \\
LLaMA1-7B & CoT    & 0 & {\color{colorknowledge}4.9\% (7)}\textsuperscript{$\dagger$}    & {\color{nicered}-34.0\%}  & {\color{nicered}4.3\% (13)}\textsuperscript{$\dagger$}                                   & {\color{nicegreen}+4.3\%}  & 4.5\%  \\
LLaMA1-7B & CoT    & 3 & {\color{colorknowledge}40.3\% (58)}  & {\color{nicegreen}+1.4\%} & {\color{nicered}31.2\% (94)}                                  & {\color{nicegreen}+31.2\%} & 34.2\%  \\

Alpaca-7B & Direct & 0 & {\color{colorknowledge}50.7\% (73)} &                          & {\color{nicered}27.6\% (83)}   &                           & 35.1\%  \\
Alpaca-7B & Direct & 3 & {\color{colorknowledge}41.0\% (59)} & {\color{nicered}-9.7\%}  & {\color{nicered}28.6\% (86)}   & {\color{nicegreen}+1.0\%} & 32.6\%  \\
Alpaca-7B & CoT    & 0 & {\color{colorknowledge}13.9\% (20)}\textsuperscript{$\ddagger$} & {\color{nicered}-36.8\%} & {\color{nicered}7.6\% (23)}\textsuperscript{$\ddagger$}    & {\color{nicered}-20.0\%}  & 9.6\%  \\
Alpaca-7B & CoT    & 3 & {\color{colorknowledge}39.6\% (57)} & {\color{nicered}-11.1\%} & {\color{nicered}33.2\% (100)}  & {\color{nicegreen}+5.6\%} & 35.3\%  \\

Vicuna-7B & Direct & 0 & {\color{colorknowledge}50.0\% (72)}                          &                            & {\color{nicered}18.6\% (56)}                          &                            & 28.8\%  \\
Vicuna-7B & Direct & 3 & {\color{colorknowledge}0.7\% (1)}\textsuperscript{$\dagger$} & {\color{nicered}-49.3\%}   & {\color{nicered}1.0\% (3)}\textsuperscript{$\dagger$} & {\color{nicered}-17.6\%}   & 0.9\%  \\
Vicuna-7B & CoT    & 0 & {\color{colorknowledge}57.6\% (83)}                          & {\color{nicegreen}+7.6\%}  & {\color{nicered}22.3\% (67)}                          & {\color{nicegreen}+3.7\%}  & 33.7\% \\
Vicuna-7B & CoT    & 3 & {\color{colorknowledge}60.4\% (87)}                          & {\color{nicegreen}+10.4\%} & {\color{nicered}33.2\% (100)}                         & {\color{nicegreen}+14.6\%} & 42.0\%  \\
\hline
\addlinespace[0.1cm]
LLaMA2-7B & Direct & 0 & {\color{colorknowledge}38.2\% (55)} &                            & {\color{nicered}15.3\% (46)}       &                            & 22.7\%  \\
LLaMA2-7B & Direct & 3 & {\color{colorknowledge}45.1\% (65)} & {\color{nicegreen}+6.9\%}  & {\color{nicered}13.6\% (41)}       & {\color{nicered}-1.7\%}    & 23.8\%  \\
LLaMA2-7B & CoT    & 0 & {\color{colorknowledge}11.1\% (16)}\textsuperscript{$\star$} & {\color{nicered}-27.1\%}   & {\color{nicered}4.0\% (13)}\textsuperscript{$\star$} & {\color{nicered}-11.3\%}   & 6.5\%  \\
LLaMA2-7B & CoT    & 3 & {\color{colorknowledge}68.1\% (98)} & {\color{nicegreen}+29.9\%} & \#1 {\color{nicered}36.2\% (109)}  & {\color{nicegreen}+20.9\%} & 46.5\%  \\

LLaMA2-7B-Chat & Direct & 0 & {\color{colorknowledge}72.2\% (104)}     &                            & {\color{nicered}31.2\% (94)}  &                           & 44.5\%  \\
LLaMA2-7B-Chat & Direct & 3 & {\color{colorknowledge}70.8\% (102)}     & {\color{nicered}-1.4\%}    & {\color{nicered}31.0\% (93)}  & {\color{nicered}-0.2\%}   & 43.8\%  \\
LLaMA2-7B-Chat & CoT    & 0 & \#1 {\color{colorknowledge}86.1\% (124)} & {\color{nicegreen}+13.9\%} & {\color{nicered}28.6\% (86)}  & {\color{nicered}-2.6\%}   & 47.2\%  \\
LLaMA2-7B-Chat & CoT    & 3 & {\color{colorknowledge}77.1\% (111)}     & {\color{nicegreen}+4.9\%}  & {\color{nicered}33.2\% (100)} & {\color{nicegreen}+2.0\%} & 47.4\%  \\
\hline
\addlinespace[0.1cm]
Pythia-7B & Direct & 0 & {\color{colorknowledge}2.8\% (4)}\textsuperscript{$\star$} &                            & {\color{nicered}18.9\% (57)} &                           & 13.7\%  \\
Pythia-7B & Direct & 3 & {\color{colorknowledge}15.3\% (22)}                       & {\color{nicegreen}+12.5\%} & {\color{nicered}15.3\% (46)} & {\color{nicered}-3.6\%}   & 15.3\%  \\
Pythia-7B & CoT    & 0 & {\color{colorknowledge}8.3\% (12)}                        & {\color{nicegreen}+5.5\%}  & {\color{nicered}12.0\% (36)} & {\color{nicered}-6.9\%}   & 10.8\%  \\
Pythia-7B & CoT    & 3 & {\color{colorknowledge}14.6\% (21)}                       & {\color{nicegreen}+11.8\%} & {\color{nicered}26.9\% (81)} & {\color{nicegreen}+8.0\%} & 22.9\%  \\
\hline
\addlinespace[0.1cm]
Flan-T5-XXL-11B & Direct & 0 & {\color{colorknowledge}88.2\% (127)}     &                           &     {\color{nicered}46.2\% (139)} &                           & 59.8\%  \\
Flan-T5-XXL-11B & Direct & 3 & \#1 {\color{colorknowledge}88.9\% (128)} & {\color{nicegreen}+0.7\%} & \#1 {\color{nicered}47.8\% (144)} & {\color{nicegreen}+1.6\%} & 61.1\%  \\
Flan-T5-XXL-11B & CoT    & 0 & {\color{colorknowledge}79.2\% (114)}     & {\color{nicegreen}-9.0\%} &     {\color{nicered}37.9\% (114)} & {\color{nicered}-8.3\%}   & 51.2\%  \\
Flan-T5-XXL-11B & CoT    & 3 & {\color{colorknowledge}86.1\% (124)}     & {\color{nicegreen}-2.1\%} &     {\color{nicered}40.5\% (122)} & {\color{nicered}-5.7\%}   & 55.3\%  \\
\hline
\addlinespace[0.1cm]
Pythia-12B & Direct & 0 & {\color{colorknowledge}20.1\% (29)} &                           & {\color{nicered}24.9\% (75)} &                           & 23.4\%  \\
Pythia-12B & Direct & 3 & {\color{colorknowledge}20.1\% (29)} & {\color{nicegreen}+0.0\%} & {\color{nicered}26.6\% (80)} & {\color{nicegreen}+1.7\%} & 24.5\%  \\
Pythia-12B & CoT    & 0 & {\color{colorknowledge}18.1\% (26)} & {\color{nicered}-2.0\%}   & {\color{nicered}19.3\% (58)} & {\color{nicered}-5.6\%}   & 18.9\%  \\
Pythia-12B & CoT    & 3 & {\color{colorknowledge}22.2\% (32)} & {\color{nicegreen}+2.1\%} & {\color{nicered}22.3\% (67)} & {\color{nicered}-2.6\%}   & 22.2\%  \\
\hline
\addlinespace[0.1cm]
LLaMA1-13B & Direct & 0 & {\color{colorknowledge}33.3\% (48)} &                            & {\color{nicered}7.6\% (23)}   &                            & 15.9\%  \\
LLaMA1-13B & Direct & 3 & {\color{colorknowledge}66.7\% (96)} & {\color{nicegreen}+33.4\%} & {\color{nicered}26.2\% (79)}  & {\color{nicegreen}+18.6\%} & 39.3\%  \\
LLaMA1-13B & CoT    & 0 & {\color{colorknowledge}25.0\% (36)} & {\color{nicered}-8.3\%}    & {\color{nicered}11.3\% (34)}  & {\color{nicegreen}+3.7\%}  & 15.7\%  \\
LLaMA1-13B & CoT    & 3 & {\color{colorknowledge}65.3\% (94)} & {\color{nicegreen}+32.0\%} & {\color{nicered}33.6\% (101)} & {\color{nicegreen}+26.0\%} & 43.8\%  \\
\hline
\addlinespace[0.1cm]
LLaMA2-13B & Direct & 0 & {\color{colorknowledge}72.2\% (104)} &                            & {\color{nicered}30.2\% (91)}  &                            & 43.8\%  \\
LLaMA2-13B & Direct & 3 & {\color{colorknowledge}81.3\% (117)} & {\color{nicegreen}+9.1\%}  & {\color{nicered}16.3\% (49)}  & {\color{nicered}-9.3\%}    & 37.3\%  \\
LLaMA2-13B & CoT    & 0 & {\color{colorknowledge}25.7\% (37)}  & {\color{nicered}+46.5\%}   & {\color{nicered}10.6\% (32)}  & {\color{nicered}-19.6\%}   & 15.5\%  \\
LLaMA2-13B & CoT    & 3 & {\color{colorknowledge}83.3\% (120)} & {\color{nicegreen}+11.1\%} & {\color{nicered}42.2\% (127)} & {\color{nicegreen}+19.9\%} & 55.5\%  \\

LLaMA2-13B-Chat & Direct & 0 & {\color{colorknowledge}88.2\% (127)} &                          & {\color{nicered}35.2\% (106)} &                           & 52.4\%  \\
LLaMA2-13B-Chat & Direct & 3 & {\color{colorknowledge}74.3\% (107)} & {\color{nicered}-13.9\%} & {\color{nicered}36.2\% (109)} & {\color{nicered}-1.0\%}   & 48.5\%  \\
LLaMA2-13B-Chat & CoT    & 0 & {\color{colorknowledge}84.7\% (122)} & {\color{nicered}-3.5\%}  & {\color{nicered}38.9\% (117)} & {\color{nicegreen}+3.7\%} & 53.7\%  \\
LLaMA2-13B-Chat & CoT    & 3 & {\color{colorknowledge}85.4\% (123)} & {\color{nicered}-2.8\%}  & {\color{nicered}40.2\% (121)} & {\color{nicegreen}+5.0\%} & 54.8\%  \\
\hline
\addlinespace[0.1cm]
Vicuna-13B & Direct & 0 & {\color{colorknowledge}63.2\% (91)}  &                            & {\color{nicered}19.9\% (60)}  &                           & 33.9\%  \\
Vicuna-13B & Direct & 3 & {\color{colorknowledge}6.9\% (10)}\textsuperscript{$\ddagger$}   & {\color{nicered}-56.3\%}   & {\color{nicered}10.6\% (32)}  & {\color{nicered}-9.3\%}   & 9.4\%  \\
Vicuna-13B & CoT    & 0 & {\color{colorknowledge}72.9\% (105)} & {\color{nicegreen}+9.7\%}  & {\color{nicered}30.6\% (92)}  & {\color{nicegreen}+10.7\%}& 44.3\%  \\
Vicuna-13B & CoT    & 3 & {\color{colorknowledge}83.3\% (120)} & {\color{nicegreen}+20.1\%} & {\color{nicered}38.9\% (117)} & {\color{nicegreen}+9.0\%} & 53.3\%  \\
\hline
\addlinespace[0.1cm]
Flan-UL2-20B & Direct & 0 & {\color{colorknowledge}87.5\% (126)}     &                           & {\color{nicered}41.9\% (126)}     & & 56.6\%  \\
Flan-UL2-20B & Direct & 3 & {\color{colorknowledge}88.9\% (128)}     & {\color{nicegreen}+1.4\%} & \#1 {\color{nicered}44.5\% (134)} & {\color{nicegreen}+2.6\%} & 58.9\%  \\
Flan-UL2-20B& CoT & 0     & {\color{colorknowledge}86.8\% (125)}     & {\color{nicered}-0.7\%}   & {\color{nicered}38.9\% (117)}     & {\color{nicegreen}-3.0\%} & 54.4\%  \\
Flan-UL2-20B& CoT & 3     & \#1 {\color{colorknowledge}88.9\% (128)} & {\color{nicegreen}+1.4\%} & {\color{nicered}38.5\% (116)}     & {\color{nicegreen}-3.4\%} & 54.8\%  \\
\hline
\addlinespace[0.1cm]
LLaMA1-30B  & Direct & 0 & {\color{colorknowledge}78.5\% (113)} &                           & {\color{nicered} 0.7\% (2)}\textsuperscript{$\star$}   &                            & 25.8\%  \\
LLaMA1-30B  & Direct & 3 & {\color{colorknowledge}86.8\% (125)} & {\color{nicegreen}+8.3\%} & {\color{nicered}38.5\% (116)} & {\color{nicegreen}+37.8\%} & 54.2\%  \\
LLaMA1-30B  & CoT & 0    & {\color{colorknowledge}25.7\% (37)}  & {\color{nicered}-52.8\%}  & {\color{nicered}13.6\% (41)}  & {\color{nicegreen}+18.9\%} & 17.5\%  \\
LLaMA1-30B  & CoT & 3    & {\color{colorknowledge}82.0\% (118)} & {\color{nicegreen}+3.5\%} & {\color{nicered}41.2\% (124)} & {\color{nicegreen}+40.5\%} & 54.4\%  \\
\hline
\addlinespace[0.1cm]
Vicuna-33B  & Direct & 0 & {\color{colorknowledge}71.5\% (103)}  &                            & {\color{nicered}36.2\% (109)} &                           & 47.6\%  \\
Vicuna-33B  & Direct & 3 & {\color{colorknowledge}81.9\% (118)}  & {\color{nicegreen}+10.4\%} & {\color{nicered}37.9\% (114)} & {\color{nicegreen}+1.7\%} & 52.1\%  \\
Vicuna-33B  & CoT & 0    & {\color{colorknowledge}74.3\% (107)}  & {\color{nicegreen}+2.8\%}  & {\color{nicered}36.5\% (110)} & {\color{nicegreen}+0.3\%} & 48.8\%  \\
Vicuna-33B  & CoT & 3    & {\color{colorknowledge}77.8\% (112)}  & {\color{nicegreen}+6.3\%}  & {\color{nicered}42.5\% (128)} & {\color{nicegreen}+6.3\%} & 53.9\%  \\
\hline
\addlinespace[0.1cm]
LLaMA1-65B  & Direct & 0 & {\color{colorknowledge}0.0\% (0)}\textsuperscript{$\star$}    &                            & {\color{nicered}17.3\% (52)}  &                             & 11.7\%  \\
LLaMA1-65B  & Direct & 3 & {\color{colorknowledge}86.1\% (124)} & {\color{nicegreen}+86.1\%} & {\color{nicered}42.2\% (127)} & {\color{nicegreen}+24.9\%}  & 56.4\%  \\
LLaMA1-65B  & CoT    & 0 & {\color{colorknowledge}23.6\% (34)}  & {\color{nicegreen}+23.6\%} & {\color{nicered}9.0\%  (27)}\textsuperscript{$\star$}  & {\color{nicered} -8.3\%}    & 13.7\%  \\
LLaMA1-65B  & CoT    & 3 & {\color{colorknowledge}86.8\% (125)} & {\color{nicegreen}+86.8\%} & {\color{nicered}47.8\% (144)} & {\color{nicegreen}+30.5\%}  & 60.4\%  \\

LLaMA2-70B  & Direct & 0 &     {\color{colorknowledge}84.7\% (122)} &                           &     {\color{nicered} 5.0\% (15)}\textsuperscript{$\star$} &    & 30.8\%  \\
LLaMA2-70B  & Direct & 3 & \#1 {\color{colorknowledge}91.6\% (132)} & {\color{nicegreen}+6.9\%} & \#1 {\color{nicered}50.8\% (153)} & {\color{nicegreen}+45.8\%} & 64.0\%  \\
LLaMA2-70B  & CoT    & 0 &     {\color{colorknowledge}36.8\% (53)}  & {\color{nicered}-47.9\%}  &     {\color{nicered}12.6\% (38)}  & {\color{nicegreen} +7.6\%} & 20.4\%  \\
LLaMA2-70B  & CoT    & 3 &     {\color{colorknowledge}85.4\% (123)} & {\color{nicegreen}+0.7\%} &     {\color{nicered}50.2\% (151)} & {\color{nicegreen}+45.2\%} & 61.6\%  \\

LLaMA2-70B-Chat  & Direct & 0 & {\color{colorknowledge}86.8\% (125)} &                           & {\color{nicered}42.2\% (127)}  &                           & 56.6\%  \\
LLaMA2-70B-Chat  & Direct & 3 & {\color{colorknowledge}88.9\% (128)} & {\color{nicegreen}+2.1\%} & {\color{nicered}42.2\% (127)}  & {\color{nicegreen}+0.0\%} & 57.3\%  \\
LLaMA2-70B-Chat  & CoT & 0    & {\color{colorknowledge}88.2\% (127)} & {\color{nicegreen}+1.4\%} & {\color{nicered}44.2\% (133)}  & {\color{nicegreen}+2.0\%} & 58.4\%  \\
LLaMA2-70B-Chat  & CoT & 3    & {\color{colorknowledge}88.2\% (127)} & {\color{nicegreen}+1.4\%} & {\color{nicered}48.5\% (146)}  & {\color{nicegreen}+6.3\%} & 61.3\%  \\
\hline
\addlinespace[0.1cm]
GPT-3.5-turbo  & Direct & 0 & {\color{colorknowledge}93.0\% (134)} & & {\color{nicered}49.1\% (148)} &  & 63.4\%  \\
GPT-3.5-turbo  & Direct & 3 & {\color{colorknowledge}95.8\% (138)} & {\color{nicegreen}+2.8\%} & {\color{nicered}53.5\% (161)} &{\color{nicegreen}+4.4\%} & 67.2\%  \\
GPT-3.5-turbo  & CoT & 0    & {\color{colorknowledge}85.4\% (123)} & {\color{nicered}-7.6\%} & {\color{nicered}54.2\% (163)} & {\color{nicegreen}+5.1\%} & 64.3\%  \\
GPT-3.5-turbo  & CoT & 3    & {\color{colorknowledge}91.7\% (132)} & {\color{nicered}-1.3\%} & {\color{nicered}56.8\% (171)} & {\color{nicegreen}+7.7\%} & 68.1\%  \\
\hline
\addlinespace[0.1cm]
GPT-4   & Direct & 0 & {\color{colorknowledge}96.5\% (139)} & & {\color{nicered}73.4\% (221)} &  & 80.9\%  \\
GPT-4  & Direct & 3 & {\color{colorknowledge}97.2\% (140)}  & {\color{nicered}+0.7\%} & {\color{nicered}73.1\% (220)} & {\color{nicegreen}+0.3\%} & 80.9\%  \\
GPT-4  & CoT & 0    & {\color{colorknowledge}96.5\% (139)}  & {\color{nicegreen}+0.0\%} & {\color{nicered}77.4\% (233)} & {\color{nicegreen}+4.0\%} & 83.6\%  \\
GPT-4  & CoT & 3    & {\color{colorknowledge}97.2\% (140)}  & {\color{nicegreen}+0.7\%} & {\color{nicered}77.4\% (233)} & {\color{nicegreen}+4.0\%} & 83.8\%  \\
\addlinespace[0.1cm]
\bottomrule
\addlinespace[0.05cm]

\end{tabular}
  }
\end{table}

\begin{table}
  \caption{Generated Self-Consistency CoT from GPT-3.5 for three CoT prompts on a sample for the Application Questions set.}
  \label{tab:sli-sample-sc-GPT-3.5}
  \begin{center}
  \resizebox*{!}{0.85\textheight}{
  \begin{tabular}{cccccc}
\toprule
\multicolumn{6}{p{\linewidth}}{
SLIQ-LL [AQ]: What is a reasonable way to use filled pauses among the following sentences?

\textbf{A) "Emm let me think"} \newline
B) Well I er think that um the movie was uh very good \newline
C) Uh my name is er Li and I am from um Beijing \newline
D) Um I uh like er reading books
}\\
\midrule
\multicolumn{2}{p{0.3\linewidth}}{
\cmark\ \textbf{Let's think step by step.}
\hlc[COLORB!50]{Filled pauses, such as "um," "uh," "er," are natural and common in spoken language. they are often used when the speaker is pausing to gather their thoughts or to indicate hesitation.}
\hlc[COLORC!50]{At the options given, all of them contain filled pauses.}
\hlc[COLORA!50]{however, it is important to use them in a way that sounds natural in speech. option a: "emmm let me think" - this is an acceptable use of a filled pause before the statement. option b: "well i er think that um the movie was uh very good" - this contains multiple filled pauses in a row, which may sound unnatural. it would be better to use just one or two filled pauses in this sentence. option c: "uh my name is er li and i am from um beijing" - this sentence also contains multiple filled pauses in a row. it would be better to use just one or two filled pauses in this sentence as well. option d: "um i uh like er reading books" - this sentence contains multiple filled pauses in a row. it would be better to use just one or two filled pauses in this sentence.} 
Based on this analysis, option a is the most naturally sounding use of filled pauses.
\textbf{So the answer is A.}
}
& \multicolumn{2}{p{0.3\linewidth}}{
\xmark\ \textbf{Let's think step by step.}

\hlc[COLORB!50]{Filled pauses, such as "uh," "um," "er," or "well," are often used in speech as hesitation devices or to fill in gaps while the speaker is thinking or searching for words.}
\hlc[COLORD!50]{Among the given sentences, the most reasonable use of filled pauses would be in sentence b. "well i er think that um the movie was uh very good.}
\hlc[COLORE!50]{In this sentence, the filled pauses are used naturally to indicate hesitation or to fill in gaps while the speaker is expressing their thoughts about the movie.} 
\hlc[COLORF!50]{The other sentences do not naturally require the use of filled pauses or have excessive use of them, making them less appropriate in a typical conversation.}

\textbf{So the answer is B. well i er think that um the movie was uh very good.}
} 
& \multicolumn{2}{p{0.3\linewidth}}{
\cmark\ \textbf{Let's think step by step.}
\hlc[COLORB!50]{Filled pauses, such as "um," "uh," "er," and "uhm," are often used in speech to fill gaps or pauses when the speaker is thinking or hesitant. filled pauses can also convey uncertainty, lack of fluency, or a need for additional time to gather thoughts.}
Looking at the given sentences: 
a. "emm let me think" - \hlc[COLORD!50]{this sentence does not contain any filled pauses.}
b. "well i er think that um the movie was uh very good" - \hlc[COLORD!50]{this sentence uses filled pauses appropriately. the filled pauses ("er," "um," and "uh") indicate hesitancy or a need for additional time to express thoughts.}
c. "uh my name is er li and i am from um beijing" - \hlc[COLORD!50]{this sentence also uses filled pauses appropriately. the filled pauses ("uh," "er," and "um") convey hesitancy or a need for additional time to convey information.}
d. "um i uh like er reading books" - \hlc[COLORD!50]{this sentence uses filled pauses appropriately. the filled pauses ("um," "uh," and "er") indicate hesitancy or a need for additional time to express thoughts.}
\hlc[COLORA!50]{Therefore, all the sentences except for a. demonstrate appropriate use of filled pauses.}

\textbf{So the answer is A.}
}\\
\midrule
\\
\multicolumn{1}{p{0.1\linewidth}}{\hlc[COLORA!50]{Pattern A}}
&\multicolumn{2}{p{0.36\linewidth}}{\PatternA}
&\multicolumn{1}{p{0.1\linewidth}}{\hlc[COLORD!50]{Pattern D}}
&\multicolumn{2}{p{0.36\linewidth}}{\PatternD}
\\
\addlinespace[0.1cm]
\multicolumn{1}{p{0.1\linewidth}}{\hlc[COLORB!50]{Pattern B}}
&\multicolumn{2}{p{0.36\linewidth}}{\PatternB}
&\multicolumn{1}{p{0.1\linewidth}}{\hlc[COLORE!50]{Pattern E}}
&\multicolumn{2}{p{0.36\linewidth}}{\PatternE}
\\
\addlinespace[0.1cm]
\multicolumn{1}{p{0.1\linewidth}}{\hlc[COLORC!50]{Pattern C}}
&\multicolumn{2}{p{0.36\linewidth}}{\PatternC}
&\multicolumn{1}{p{0.1\linewidth}}{\hlc[COLORF!50]{Pattern F}}
&\multicolumn{2}{p{0.36\linewidth}}{\PatternF}
\\
\bottomrule
\end{tabular}
  }
  \end{center}
\end{table}

\begin{table}
  \caption{Generated Self-Consistency CoT from LLaMA2-70B-chat for three CoT prompts on a sample for the Application Questions set.}
  \label{tab:sli-sample-sc-llama1}
  \begin{center}
  \resizebox*{!}{0.85\textheight}{
  \begin{tabular}{cccccc}
\toprule
\multicolumn{6}{p{\linewidth}}{
SLIQ-LL [AQ]: His uncle sent him to school. If you read this sentence with school on the high tone, followed by a sharp fall, what is implied?

A) his uncle did not send him away from school to work at home or anywhere else - on the contrary, he sent him to school \newline
\textbf{B) he was sent to school, not to any other institution.} \newline
C) his uncle did send him to school - no doubt about that; \newline
D) I was sent to school by his uncle, not his brother or sister;
}\\
\midrule
\multicolumn{2}{p{0.3\linewidth}}{
\xmark\ \textbf{Let's think step by step.}

The correct answer is C. his uncle did send him to school - no doubt about that.
\hlc[COLORB!50]{When the word \"school\" is stressed and followed by a sharp fall in tone, it implies that the speaker is emphasizing that the person was indeed sent to school, and that there is no question or doubt about it. This tone can be used to convey a sense of certainty or confidence in the statement being made.}
The other options do not accurately convey the meaning of the sentence with the given intonation pattern: \hlc[COLORC!50]{A. his uncle did not send him away from school to work at home or anywhere else - on the contrary, he sent him to school; - This option implies a contrast between going to school and staying at home or working elsewhere,} \hlc[COLORA!50]{which is not implied by the given sentence.} \hlc[COLORC!50]{B. he was sent to school, not to any other institution. - This option implies that the person was sent to school instead of some other institution,} \hlc[COLORD!50]{which is not implied by the given sentence.} \hlc[COLORC!50]{D. I was sent to school by his uncle, not his brother or sister; - This option implies that the speaker is distinguishing between being sent to school by their uncle versus being sent by their brother or sister,} \hlc[COLORA!50]{which is not implied by the given sentence}
\textbf{So the answer is C.}
}
& \multicolumn{2}{p{0.3\linewidth}}{
\xmark\ \textbf{Let's think step by step.}

The correct answer is C. his uncle did send him to school - no doubt about that.
\hlc[COLORD!50]{When the sentence \"His uncle sent him to school\" is read with the word \"school\" on a high tone followed by a sharp fall, it implies that the speaker is emphasizing the fact that the person was sent to school by their uncle, and that there is no doubt or question about it.} \hlc[COLORB!50]{The falling tone after the high tone suggests a sense of certainty or finality, indicating that the matter is settled and there is no need for further discussion.} The other options do not accurately convey the same meaning. \hlc[COLORC!50]{Option A suggests that the person was sent to school instead of being sent away from school, which is not necessarily implied by the sentence.} \hlc[COLORF!50]{Option B simply states that the person was sent to school, without emphasizing the fact that it was the uncle who sent them.} \hlc[COLORA!50]{Option D suggests that the speaker is clarifying who sent them to school, which is not necessarily implied by the sentence.}

\textbf{So the answer is C.}
}
& \multicolumn{2}{p{0.3\linewidth}}{
\cmark\ \textbf{Let's think step by step.}

The correct answer is B. he was sent to school, not to any other institution.
\hlc[COLORA!50]{When the word \"school\" is stressed with a high tone followed by a sharp fall, it implies that the speaker is emphasizing that the person was sent to school and not to any other institution or place. The stress on \"school\" draws attention to the specific location where the person was sent, implying that it was a deliberate choice made by the uncle.}
The other options do not convey the same meaning. \hlc[COLORA!50]{Option A implies that the person was sent to school but also implies that there was a possibility of being sent elsewhere.} \hlc[COLORF!50]{Option C simply states that the person was sent to school without any additional emphasis or implication.} \hlc[COLORC!50]{Option D implies that the person was sent to school by the uncle, but also implies that there may have been other people involved in the decision.}

\textbf{So the answer is B.}
}\\
\midrule
\\
\multicolumn{1}{p{0.1\linewidth}}{\hlc[COLORA!50]{Pattern A}}
&\multicolumn{2}{p{0.36\linewidth}}{\PatternA}
&\multicolumn{1}{p{0.1\linewidth}}{\hlc[COLORD!50]{Pattern D}}
&\multicolumn{2}{p{0.36\linewidth}}{\PatternD}
\\
\addlinespace[0.1cm]
\multicolumn{1}{p{0.1\linewidth}}{\hlc[COLORB!50]{Pattern B}}
&\multicolumn{2}{p{0.36\linewidth}}{\PatternB}
&\multicolumn{1}{p{0.1\linewidth}}{\hlc[COLORE!50]{Pattern E}}
&\multicolumn{2}{p{0.36\linewidth}}{\PatternE}
\\
\addlinespace[0.1cm]
\multicolumn{1}{p{0.1\linewidth}}{\hlc[COLORC!50]{Pattern C}}
&\multicolumn{2}{p{0.36\linewidth}}{\PatternC}
&\multicolumn{1}{p{0.1\linewidth}}{\hlc[COLORF!50]{Pattern F}}
&\multicolumn{2}{p{0.36\linewidth}}{\PatternF}
\\
\bottomrule
\end{tabular}
  }
  \end{center}
\end{table}
\begin{table}
  \caption{\small A CAPT example \#1 and Promt templates for Chat Interface.}
  \label{tab:sli-sample-chat-capt}
  \centering
  \begin{tabular}{cc}
    \toprule
    \multicolumn{2}{c}{\bf System Prompt} \\
    \midrule
    \addlinespace[0.1cm]
    \multicolumn{2}{p{\linewidth}}{You are an expert in phonetics, English phonology and second language acquisition. You will play the role of an English teacher who is helping me practice American English pronunciation and speaking skills. You will see a spoken speech evaluation result, where \textbf{context} provides the context for this pronunciation, \textbf{canonical} is the text of this evaluation and represents the expected pronunciation for the speaker, \textbf{soundlike} is how the user's pronunciation actually sounds like, SentenceScore is the sentence pronunciation score, a higher score for better pronunciation. \textbf{Fluency} is the sentence fluency score, a higher score for better fluency. \textbf{Speed} is the speech rate, which is the average number of milliseconds per phoneme, \textbf{emotion} is the emotion. In \textbf{WordScores}, each word score is shown in parentheses, \textbf{PhonesScores} contains the phone pronunciation score for each word and what these phones actually sound like. \textbf{Liaison} represents the connected sounds between two words, marked with a [$\sim$] symbol. \textbf{Break} represents a break between two words that is greater than 200ms, marked with a [pause] symbol. \textbf{Stress} represents the emphasized syllable or word in a sentence, marked with an [*] symbol to indicate a word that is more emphasized than the others being compared. \textbf{Intonation} indicates whether the sentence's intonation is rising or falling, or flat.
    } \\
    \addlinespace[0.1cm]
    \toprule
    \multicolumn{2}{c}{\bf Example \#1} \\
    \midrule
    \addlinespace[0.1cm]
    \multicolumn{2}{p{\linewidth}}{
    <Context>: an educational quiz or test about seasons \newline
    <Canonical>: He did this every day for a month, but he never found anything.\newline
    <Soundlike>: He did this every day for a month, but he never found anything.\newline
    <SentenceScore-Fluency-Speed-emotion>: 100 100 121 0\newline
    <WordScores>:He(98) did(100) this(100) every(100) day(99) for(83) a(91) month(92) but(99) he(100) never(87) found(92) anything(100)\newline
    <PhonesScores>: \newline
    \quad * He: h(86) \textipa{i:}(100); \newline
    \quad    * did: d(100) \textipa{I}(100) d(100); \newline
    \quad    * this: \textipa{D}(100) \textipa{I}(100) s(100); \newline
    \quad    * every: e(76) v(100) r(100) i(100); \newline
    \quad    * day: d(100) e\textipa{I}(96); \newline
    \quad    * for: f(97) \textipa{O:}r(63); \newline
    \quad    * a: \textipa{@}(91); \newline
    \quad    * month: m(78) \textipa{2}(100) n(100) \textipa{T}(83); Among them, /\textipa{T}/ was mispronounced as /z/. \newline
    \quad    * but: b(71) \textipa{2}(100) t(100); \newline
    \quad    * he: h(100) \textipa{i:}(100); \newline
    \quad    * never: n(83) e(100) v(100) \textipa{@}r(53); \newline
    \quad    * found: f(72) a\textipa{U}(83) n(96) d(100); \newline
    \quad    * anything: e(68) n(73) i(100) \textipa{T}(100) \textipa{I}(100) \textipa{N}(100); Among them, /e/ was mispronounced as /\textipa{@}/ \newline
    <Liaison>: He did~this every day for$\sim$a month but$\sim$he never found$\sim$anything \newline
    <Break>: He did this every day for a month [pause] but he never found anythingg \newline
    <Stress>: He did this every day for a month but he never found anything \newline
    <Intonation>: He did this every day for<rise> a month but he never found<fall> anything
    } \\
    \addlinespace[0.1cm]
    \toprule
    \multicolumn{2}{c}{\bf Chat Prompts} \\
    \midrule
    \addlinespace[0.1cm]
    \multicolumn{2}{p{\linewidth}}{
    User:{\color{niceblue}\texttt{[Example]}} \newline
    \phantom{User:}please tell me whether the intonation conforms to the rules and habits of English pronunciation and expression. \newline
    Bot: {\color{nicered}\texttt{<completion>}}  \newline
    User: please tell me whether the pronunciation is clear, accurate and natural. \newline
    Bot: {\color{nicered}\texttt{<completion>}}   \newline
    User: Please tell me how to pronounce it correctly. \newline
    Bot:  {\color{nicered}\texttt{<completion>}}   \newline
    User: please tell me whether the grammar is correct, fluent and standard. \newline
    Bot:  {\color{nicered}\texttt{<completion>}}   \newline
    User: Please provide an example to correct my mistake. \newline
    Bot:  {\color{nicered}\texttt{<completion>}}   \newline
    User: please tell me whether the communication is effective, positive and polite. \newline
    Bot:  {\color{nicered}\texttt{<completion>}}   \newline
    User: Please provide two examples to correct my mistake. \newline
    Bot:  {\color{nicered}\texttt{<completion>}} 
    } \\
    \bottomrule
\end{tabular}
\end{table}

\begin{table}
  \caption{\small CAPT example \#2.}
  \label{tab:sli-sample-chat-capt2}
  \centering
  \begin{tabular}{cc}
    \toprule
    \multicolumn{2}{c}{\bf Example \#2} \\
    \midrule
    \addlinespace[0.1cm]
    \multicolumn{2}{p{\linewidth}}{
    <Context>: a children’s alphabet book \newline
    <Canonical>: B is for bounce. \newline
    <Soundlike>: This for best. \newline
    <SentenceScore-Fluency-Speed-emotion>: 48 55 131 0 \newline
    <WordScores>:B(62) is(73) for(54) bounce(23) \newline
    <PhonesScores>: \newline
     * B: b(65) \textipa{i:}(59); \newline
     * is: \textipa{I}(53) z(87); \newline
     * for: f(57) \textipa{O:}r(28); Among them, /f/ was mispronounced as /w/. \newline
     * bounce: b(76) a\textipa{U}(0) n(18) s(0); \newline
    <Liaison>: B~is for bounce  \newline
    <Break>: B is for bounce  \newline
    <Stress>: B is for bounce  \newline
    <Intonation>: B is for bounce 
    }\\
    \bottomrule

\end{tabular}
\end{table}

\begin{table}
  \caption{\small A multi-turn conversation example of GPT-3.5 performed in Example \#1.}
  \label{tab:slill-sample-capt-GPT-3.5}
  \centering
  \resizebox{0.85\linewidth}{!}{
  \begin{tabular}{cc}
    \toprule
    \multicolumn{2}{c}{\bf System Prompt} \\
    \midrule
    \addlinespace[0.1cm]
    \multicolumn{2}{p{\linewidth}}{
    \textbf{User:}{\color{niceblue}\texttt{[Example \#1]}}~\ref{tab:sli-sample-chat-capt}. \textbf{please tell me whether the intonation conforms to the rules and habits of English pronunciation and expression.} \newline
    \hlc[AA!40]{[RATING-A]} GPT3.5-turbo: based on the evaluation, the intonation in the sentence “he did this every day for a month but he never found anything“ conforms to the rules and habits of english pronunciation and expression. the sentence has a rising intonation on “for“ and a falling intonation on “found,“ which is appropriate for the meaning and structure of the sentence. \newline
    \textbf{User: please tell me whether the pronunciation is clear, accurate and natural.} \newline
    \hlc[AA!40]{[RATING-A]} GPT3.5-turbo: based on the evaluation, the pronunciation is clear, accurate, and natural. the speaker's pronunciation of the words and sounds is generally accurate, with only a few minor mispronunciations. the sentence is spoken clearly and fluently, with a natural pace and rhythm. overall, the pronunciation is effective and should be easily understood by listeners. \newline
    \textbf{User: Please tell me how to pronounce it correctly.} \newline
    \hlc[BB!40]{[RATING-B]} GPT3.5-turbo:  to improve the pronunciation, here are some specific suggestions for the mispronounced words: - “month“: the /\textipa{T}/ sound in “month“ should be pronounced as a voiceless “th“ sound, like in “think“ or “thought,“ rather than a voiced “th“ sound like in “this“ or “that.“ - “anything“: the vowel sound in “anything“ should be /\textipa{I}/ as in “sit“ or “bit,“ instead of /\textipa{@}/ as in “about“ or “above.“ additionally, pay attention to the stress patterns in the sentence. emphasize the stressed syllables or words, such as “month“ and “found,“ to help convey the meaning effectively. practice these corrections by listening to native speakers, repeating the words and sentences aloud, and using resources like online pronunciation guides or language-learning apps. regular practice and feedback will help you improve your pronunciation and sound more natural in your spoken english. \newline
    \textbf{User: please tell me whether the grammar is correct, fluent and standard.} \newline
    \hlc[AA!40]{[RATING-A]} GPT3.5-turbo:  based on the given sentence “he did this every day for a month, but he never found anything,“ the grammar is correct, fluent, and standard. the sentence follows standard english grammar rules and is structurally sound. the tenses are used appropriately, and there are no noticeable grammar errors or inconsistencies. the sentence flows smoothly and naturally, conveying the intended meaning effectively. overall, the grammar in the sentence is accurate, fluent, and in line with standard english usage. \newline
    \textbf{User: Please provide an example to correct my mistake.} \newline
    \hlc[BB!40]{[RATING-B]}  GPT3.5-turbo:  certainly! here's an example to further illustrate the correct grammar, fluency, and standard usage: original sentence: “he did this every day for a month, but he never found anything.“ explanation: in this sentence, the subject “he“ is followed by the past tense verb “did,“ indicating an action that occurred in the past. the adverb “every day“ describes the frequency of the action. the preposition “for“ is used to indicate the duration of time, followed by the noun phrase “a month.“ the coordinating conjunction “but“ connects the two independent clauses in the sentence. the verb “found“ is in the past tense, and the pronoun “anything“ functions as the object of the verb. example: “she studied english diligently for a year, but she never achieved fluency.“ explanation: in this example, the subject “she“ is followed by the past tense verb “studied,“ indicating a past action. the adverb “diligently“ describes how she studied. the preposition “for“ is used to indicate the duration of time, followed by the noun phrase “a year.“ the coordinating conjunction “but“ connects the two independent clauses. the verb “achieved“ is in the past tense, and the noun “fluency“ functions as the object of the verb. in both the original sentence and the example, the grammar is correct, the sentence flows smoothly, and the usage is standard in english. \newline
    \textbf{User: please tell me whether the communication is effective, positive and polite.} \newline
    \hlc[AA!40]{[RATING-A]} GPT3.5-turbo:  based on the information provided, i can say that the communication is effective, positive, and polite. the speaker's tone is respectful and helpful, providing constructive feedback and suggestions for improvement. the language used is clear and concise, ensuring that the message is easily understood. the evaluation is objective and focuses on the strengths and areas for improvement without being overly critical or negative. overall, the communication is effective in conveying the necessary information in a positive and polite manner. \newline
    \textbf{User: Please provide two examples to correct my mistake.} \newline
    \hlc[BB!40]{[RATING-B]} GPT3.5-turbo: certainly! here are two examples to further illustrate effective, positive, and polite communication: example 1: original: “your pronunciation needs improvement. it's not good enough.“ effective, positive, and polite: “your pronunciation is clear, and with a little more practice, you can make it even better.“ explanation: in the original statement, the feedback is direct and critical. however, in the effective, positive, and polite version, the focus is on acknowledging the existing strengths “your pronunciation is clear“ and offering encouragement for improvement “with a little more practice, you can make it even better“. this approach maintains a positive and supportive tone while still addressing the need for improvement. example 2: original: “your grammar is wrong. you made several mistakes.“ effective, positive, and polite: “your grammar is generally accurate. a few minor adjustments will help you further enhance your language skills.“ explanation: the original statement is direct and highlights the mistakes without offering any positive feedback. in the effective, positive, and polite version, the focus is on acknowledging the overall accuracy of the grammar “your grammar is generally accurate“ and offering constructive feedback “a few minor adjustments will help you further enhance your language skills. this approach encourages improvement while maintaining a positive and supportive tone. in both examples, the effective, positive, and polite communication approach emphasizes the existing strengths, provides specific areas for improvement, and offers encouragement and support to the learner. 
    } \\
    \bottomrule
\end{tabular}
  }
\end{table}

\begin{table}
  \scriptsize
  \caption{\small A multi-turn conversation example of LLaMA2-70B-Chat performed in Example \#1.}
  \label{tab:slill-sample-capt-llama}
  \centering
  \begin{tabular}{cc}
    \toprule
    \multicolumn{2}{c}{\bf System Prompt} \\
    \midrule
    \addlinespace[0.1cm]
    \multicolumn{2}{p{\linewidth}}{
    \textbf{User:}{\color{niceblue}\texttt{[Example \#1]}}~\ref{tab:sli-sample-chat-capt}. \textbf{please tell me whether the intonation conforms to the rules and habits of English pronunciation and expression.}  \newline
    \hlc[CC!40]{[RATING-C]} LLaMa2-70B-chat: The intonation of the sentence does not conform to the rules and habits of English pronunciation and expression. In English, the intonation of a sentence typically follows a pattern of rising and falling pitch. The sentence "He did this every day for a month but he never found anything" should have a rising intonation on the phrase "He did this every day for a month" and a falling intonation on the phrase "but he never found anything". However, in the given sentence, the intonation is rising on both phrases, which is not typical in English. Additionally, the rise in pitch on the first phrase is quite steep, which may give the impression that the speaker is asking a question or expressing surprise, rather than making a statement. Furthermore, the falling intonation on the second phrase is not very pronounced, which may make the sentence sound less emphatic or conclusive than it should be. To improve the intonation, the speaker could try to make the rising intonation on the first phrase less steep and more gradual, and to emphasize the falling intonation on the second phrase more, to make the sentence sound more natural and emphatic. \newline
    \textbf{User: please tell me whether the pronunciation is clear, accurate and natural.}  \newline
    \hlc[BB!40]{[RATING-B]}  LLaMa2-70B-chat: I'm happy to help! Here's my evaluation of the pronunciation in the sentence you provided: * The pronunciation of "He did this every day for a month" is generally clear and accurate. The speaker's production of the individual sounds and syllables is good, and the words are well-articulated. * However, there are a few aspects of the pronunciation that could be improved upon: + The vowel sound in "month" is pronounced a bit too open, almost like "mownth" instead of "month". To improve this, the speaker could try to make the vowel sound a bit more closed, like "muh-nth". + The consonant sound in "but" is pronounced a bit too voiced, almost like "buh-ut" instead of "but". To improve this, the speaker could try to make the consonant sound more voiceless, like "bat". + The stress on the sentence is generally good, but the speaker could try to put a bit more emphasis on the word "month" to make the sentence sound more natural. * The pronunciation of "but he never found anything" is generally natural and well-pronounced. The speaker's production of the individual sounds and syllables is good, and the words are well-articulated. * The intonation of the sentence is generally rising, which is appropriate for a sentence that is asking a question. However, the rise in pitch could be a bit more gradual, especially on the first phrase "He did this every day for a month". Overall, the pronunciation in the sentence you provided is generally clear and accurate, but there are a few aspects that could be improved upon to make it sound more natural and polished.  \newline
    \textbf{User: Please tell me how to pronounce it correctly.}  \newline
    \hlc[BB!40]{[RATING-B]} LLaMa2-70B-chat: Sure, I'd be happy to help! The word "month" can be a bit tricky to pronounce for non-native speakers, but here's how to do it correctly:  1. Start by saying the letter "m" as in "map" or "mat". Make sure your lips are together and your tongue is touching the roof of your mouth just behind your front teeth. 2. Next, move on to the "o" sound. For this sound, your tongue should be in the same position as it was for the "m" sound, but you should round your lips a bit more. Try to make your mouth shape like a small "o" shape. 3. Now, say the letter "n" as in "sing" or "sink". Your tongue should move back a bit from the roof of your mouth, and your lips should be relaxed. 4. Finally, say the letter "th" as in "this" or "thin". For this sound, your tongue should be between your teeth, and your lips should be relaxed.  Putting it all together, "month" should sound like "muh-nth". Make sure to pay attention to the stress on the first syllable (muh) and the slight dip in pitch on the second syllable (nth).  Here's a breakdown of the phonetic transcription of "month":  /m/ (the letter "m" as in "map") /u/ (the letter "o" as in "hot") /n/ (the letter "n" as in "sing") /\textipa{T}/ (the letter "th" as in "this")  I hope this helps! Remember to practice saying "month" a few times to get a feel for the correct pronunciation. Good luck! \newline
    \textbf{User: please tell me whether the grammar is correct, fluent and standard.}  \newline
    \hlc[AA!40]{[RATING-A]}  LLaMa2-70B-chat:  Sure, I'd be happy to help! Here's my evaluation of the grammar, fluency, and standardness of the sentence you provided:  Grammar:  * The sentence is grammatically correct. It has a clear subject-verb-object structure, with "He" as the subject, "did" as the verb, and "this" as the object. * The use of the present tense  "does" is appropriate for a general truth or a habitual action. * The sentence does not contain any grammatical errors, such as subject-verb agreement issues or incorrect word order.  Fluency:  * The sentence flows well and is easy to understand. * The use of the present tense  "does" helps to convey a sense of routine or habit, which is appropriate for a sentence about something that happens every day. * The sentence is concise and to the point, without any unnecessary words or phrases that might disrupt the flow.  Standardness:  * The sentence uses standard English vocabulary and grammar. * The sentence does not contain any nonstandard or regional dialect features that might make it difficult to understand for native English speakers. * The sentence is written in a neutral style, without any overly formal or informal language that might make it sound out of place in different contexts.  Overall, the sentence you provided is grammatically correct, fluent, and standard. It effectively conveys a simple statement about a person's daily routine in a clear and concise way. \newline
    \textbf{User: Please provide an example to correct my mistake.}  \newline
    \hlc[AA!40]{[RATING-A]} LLaMa2-70B-chat: Sure, here's an example to illustrate the difference between a grammatically correct sentence and a grammatically correct sentence that also sounds natural and fluent:Grammatically correct sentence:"The dog chased the cat, but it got away."This sentence is grammatically correct because it follows the rules of English grammar. It has a subject (the dog), a verb (chased), an object (the cat), and a sentence structure (SVO) that is commonly used in English.However, this sentence may not sound very natural or fluent to native English speakers, especially in informal contexts. Here's why:* The sentence is quite formal, and the use of "but" to connect the two clauses can sound a bit stilted in everyday conversation.* The verb tense is inconsistent. The sentence starts in the past tense ("the dog chased") but then switches to the present tense ("it got away"). This can make the sentence sound a bit disjointed.Now, here's a revised version of the sentence that is both grammatically correct and sounds more natural and fluent:Revised sentence:"The dog chased the cat, but it managed to escape."In this revised sentence, we've kept the basic structure of the sentence the same (SVO), but we've made a few changes to make it sound more natural and fluent:* We've replaced "but" with "but it managed to escape," which is a more common way to connect two clauses in informal speech.* We've changed the verb tense to be consistent throughout the sentence. Both clauses are now in the past tense, which makes the sentence sound more cohesive.This revised sentence still conveys the same meaning as the original sentence, but it sounds more natural and fluent, and could easily be used in everyday conversation. \newline
    \textbf{User: please tell me whether the communication is effective, positive and polite.}  \newline
    \hlc[CC!40]{[RATING-C]}  LLaMa2-70B-chat: The communication in the revised sentence is effective, positive, and polite. Effective: The revised sentence effectively conveys the intended meaning of the original sentence while also addressing the issues of formality and verb tense consistency.Positive: The revised sentence uses positive language and avoids any negative or critical tone. It focuses on the action of the dog chasing the cat, rather than the cat's ability to escape.Polite: The revised sentence is polite and respectful in tone, using phrases such as "managed to escape" instead of "got away." It also avoids any harsh or abrupt language, making it suitable for use in a professional or formal setting."  \newline
    \textbf{User: Please provide two examples to correct my mistake.}  \newline
    \hlc[AA!40]{[RATING-A]} LLaMa2-70B-chat: Sure, here are two examples to help illustrate the difference between a grammatically correct sentence and a grammatically correct sentence that also sounds natural and fluent:Example 1:Grammatically correct sentence:"The employee worked diligently for 10 hours, completing numerous tasks and exceeding expectations."Natural and fluent sentence:"The employee put in a solid 10 hours of work, knocking out task after task and exceeding expectations left and right."In this example, the first sentence is grammatically correct, but it sounds a bit stilted and formal. The second sentence, on the other hand, uses more informal language and a more natural sentence structure, making it sound more like the way a native English speaker would actually speak.Example 2:Grammatically correct sentence:"The company will initiate a new marketing campaign next quarter, utilizing various digital platforms and strategies."Natural and fluent sentence:"Next quarter, the company's launching a fresh marketing push, leveraging social media, email, and other digital channels to get the word out."In this example, the first sentence is grammatically correct, but it sounds a bit formal and jargony. The second sentence, on the other hand, uses more natural language and a more conversational tone, making it sound more like the way a native English speaker would actually speak.I hope these examples help illustrate the difference between a grammatically correct sentence and a grammatically correct sentence that also sounds natural and fluent. Let me know if you have any other questions!  
    } \\
    \bottomrule
\end{tabular}
\end{table}

\begin{table}
  \scriptsize
  \caption{\small A multi-turn conversation example of LLaMA2-70B-Chat performed in Example \#2.}
  \label{tab:slill-sample-capt-llama-f}
  \centering
  \begin{tabular}{cc}
    \toprule
    \multicolumn{2}{c}{\bf System Prompt} \\
    \midrule
    \addlinespace[0.1cm]
    \multicolumn{2}{p{\linewidth}}{
    \textbf{User:}{\color{niceblue}\texttt{[Example \#2]}}~\ref{tab:sli-sample-chat-capt2}. \textbf{please tell me whether the intonation conforms to the rules and habits of English pronunciation and expression.}   \newline
    \hlc[AA!40]{[RATING-A]} LLaMa2-70B-chat: The intonation of the sentence "B is for bounce" does not conform to the rules and habits of English pronunciation and expression. In English, the intonation pattern for a declarative sentence with a canonical structure like "B is for bounce" would typically be a falling intonation pattern, with the pitch starting at a higher level on the first word "B" and then falling on the subsequent words "is for bounce". However, in the given sentence, the intonation pattern is flat, with no clear rise or fall in pitch. Additionally, the sentence does not have a clear stress pattern, which is also a characteristic of English pronunciation and expression. In a declarative sentence like "B is for bounce", the stress would typically fall on the most important word, which in this case would be "bounce". However, the given sentence does not have a clear stress pattern, with the pitch remaining relatively flat throughout. Therefore, based on the given information, the intonation and stress patterns of the sentence "B is for bounce" do not conform to the rules and habits of English pronunciation and expression.  \newline
    \textbf{User: please tell me whether the pronunciation is clear, accurate and natural.}   \newline
    \hlc[CC!40]{[RATING-C]} LLaMa2-70B-chat: The pronunciation of the sentence "B is for bounce" in the audio clip you provided is somewhat clear, but not entirely accurate or natural. Here are some specific issues with the pronunciation: 1. The word "B" is pronounced with a voiced bilabial stop /b/ sound, which is not entirely clear or distinct. A more natural pronunciation would have a voiceless bilabial stop /p/ sound, as in "pat". 2. The vowel sound in "is" is pronounced more like a schwa sound (\textipa{@}) than the expected short "i" sound (i). This makes the word "is" sound more like "uh" than "is". 3. The word "bounce" has a slightly incorrect vowel sound. The vowel sound in "bounce" should be a long "o" sound (\textipa{O}), but in the audio clip, it sounds more like a short "o" sound (o). 4. The consonant sound in "bounce" is also not entirely clear. The "n" sound is pronounced more like a nasalized "ng" sound (\textipa{N}), which can make the word sound a bit muffled. 5. The intonation pattern of the sentence is not entirely natural. The pitch of the sentence stays relatively flat throughout, without the expected rise and fall in pitch that would occur in a natural sentence. Overall, while the pronunciation is somewhat clear, it's not entirely accurate or natural. It's possible that the speaker is still learning English and is not yet fully comfortable with the sounds and intonation patterns of the language. \newline
    \textbf{User: Please tell me how to pronounce it correctly.}   \newline
    \hlc[AA!40]{[RATING-A]} LLaMa2-70B-chat: Sure, I'd be happy to help! The word "bounce" has a long "o" sound, like "booun-suh". Here's how you can break it down: * "Bo" sounds like "bo" in "boat" * "oun" sounds like "oon" in "spoon" * "ce" sounds like "suh" in "suh-weet" Put it all together, and you get "booun-suh". Here are a few tips to help you pronounce it correctly: * Make sure to use a long "o" sound in the first syllable, like in "boat". * Pay attention to the vowel sound in the second syllable. It should be a long "o" sound, like in "spoon". * Don't forget to pronounce the "ce" sound at the end of the word. It should sound like "suh", as in "suh-weet". I hope this helps! Remember, practice makes perfect, so try saying "bounce" a few times to get a feel for the correct pronunciation. Good luck! \newline
    \textbf{User: please tell me whether the grammar is correct, fluent and standard.}   \newline
    \hlc[AA!40]{[RATING-A]} LLaMa2-70B-chat: Sure, I'd be happy to help! Here's a breakdown of the grammar, fluency, and standardness of the sentence "B is for bounce": Grammar: * The sentence is a simple sentence, with a single independent clause. * The subject of the sentence is "B", which is a singular noun. * The verb is "is", which is in the third person singular present tense. * The verb "is" agrees with the subject "B" in number and person. * The sentence does not contain any subordinate clauses or relative clauses. Overall, the grammar of the sentence "B is for bounce" is correct. Fluency: * The sentence is easy to understand and sounds natural. * The use of the letter "B" as a subject and the word "bounce" as a predicate nominate is a common convention in English. * The sentence has a clear and concise structure, with no ambiguity or confusion. Overall, the sentence "B is for bounce" flows well and sounds fluent. Standardness: * The sentence follows standard English grammar and syntax. * The vocabulary used is common and widely accepted in English. * The sentence does not contain any nonstandard or regional variations. Overall, the sentence "B is for bounce" is a standard English sentence. In conclusion, the sentence "B is for bounce" has correct grammar, sounds fluent, and follows standard English conventions. \newline
    \textbf{User: Please provide an example to correct my mistake.}   \newline
    \hlc[BB!40]{[RATING-B]}  LLaMa2-70B-chat: Sure! Here's an example to illustrate how the three components of a sentence (grammar, fluency, and standardness) work together: Example: "The dog chased the cat, but it got away." Grammar: * The sentence is a compound sentence, consisting of two independent clauses joined by a coordinating conjunction ("but"). * The subject of the first clause is "the dog", and the verb is "chased". * The subject of the second clause is "it" (referring back to "the cat"), and the verb is "got". * The sentence uses correct subject-verb agreement and word order. Fluency: * The sentence flows well and is easy to understand. * The use of a coordinating conjunction to connect the two clauses helps to create a clear and cohesive sentence. * The sentence has a natural rhythm and intonation, with a clear beginning, middle, and end. Standardness: * The sentence uses standard English vocabulary and grammar. * The sentence follows standard English word order (SVO). * The sentence does not contain any nonstandard or regional variations. Overall, this sentence is grammatically correct, flows well, and follows standard English conventions. It is a clear and effective way to convey the intended meaning. \newline
    \textbf{User: please tell me whether the communication is effective, positive and polite.}   \newline
    \hlc[DD!40]{[RATING-D]}   LLaMa2-70B-chat: Sure, I'd be happy to help you with that! The communication in the sentence "The dog chased the cat, but it got away" is effective in conveying the intended meaning, which is that the dog attempted to catch the cat but was unsuccessful. The sentence is also positive in tone, as it does not contain any negative words or phrases that might suggest a negative emotion or attitude. However, the sentence is not particularly polite, as it does not contain any phrases or words that convey politeness or respect. For example, it does not include any phrases like "I apologize" or "I appreciate your help" that might suggest politeness. Overall, the communication in this sentence is effective and positive, but it could be made more polite by adding phrases or words that convey respect or consideration for the listener or reader. \newline
    \textbf{User: Please provide two examples to correct my mistake.}   \newline
    \hlc[CC!40]{[RATING-C]} LLaMa2-70B-chat: Sure! Here are two examples to help illustrate the concept of "standardness" in English sentences: 1. Non-standard sentence: "I seen a movie last night." Standard sentence: Sure, here are two examples to help illustrate how the communication in the sentence "The dog chased the cat, but it got away" could be made more polite: 1. Adding a phrase to convey regret or apology: "I'm so sorry, but the dog chased the cat and unfortunately, it got away." In this revised sentence, the addition of the phrase "I'm so sorry" and the word "unfortunately" convey a sense of regret and apology, making the communication more polite. 2. Adding a phrase to convey appreciation or gratitude: "Thank you for letting me know that the dog chased the cat, even though it got away." In this revised sentence, the addition of the phrase "Thank you for letting me know" conveys appreciation and gratitude, making the communication more polite. I hope these examples help! Let me know if you have any further questions.  
    } \\
    \bottomrule
\end{tabular}
\end{table}

\end{document}